\documentclass{article} 
\usepackage[final]{colm2026_conference}

\usepackage{microtype}
\usepackage{hyperref}
\usepackage{booktabs}
\usepackage{latexsym}
\usepackage{url}
\usepackage{array}    
\usepackage{multirow} 
\usepackage{subcaption}
\usepackage{caption}
\usepackage{amsmath}
\usepackage{cleveref}
\usepackage{xcolor}
\usepackage{wrapfig}
\usepackage{array}
\usepackage{longtable}
\usepackage{booktabs}
\usepackage{pdflscape}
\usepackage{algorithm}
\usepackage{algpseudocode}
\usepackage{xurl}

\definecolor{cspink}{RGB}{219,48,122}
\definecolor{infoutility}{HTML}{0072B2}    
\definecolor{infoquality}{HTML}{D55E00}    
\definecolor{stylequality}{HTML}{009E73}   
\usepackage{tabularx}
\newcolumntype{Y}{>{\raggedright\arraybackslash}X}
\definecolor{infoutilityhl}{HTML}{B3D9F2}
\definecolor{infoqualityhl}{HTML}{F2C299}   
\definecolor{stylequalityhl}{HTML}{B3F2D9}  
\definecolor{cspink}{RGB}{219,48,122}

\usepackage{tcolorbox}
\usepackage{soul}
\usepackage{mdframed}
\usepackage{amssymb}
\usepackage{pifont}

\title{Measuring AI ``Slop'' in Text}

\usepackage{lineno}

\definecolor{darkblue}{rgb}{0, 0, 0.5}
\hypersetup{colorlinks=true, citecolor=darkblue, linkcolor=darkblue, urlcolor=darkblue}

\title{Measuring AI ``Slop'' in Text}

\author{\textbf{Chantal Shaib$^1$\thanks{All experimentation and data processing done at NEU.}}\quad\quad
\textbf{Tuhin Chakrabarty$^2$}\quad\quad
\textbf{Diego Garcia-Olano$^3$}\quad\quad
\textbf{Byron C. Wallace$^1$}\quad\quad \\
$^1$Northeastern University, $^2$Stony Brook University, $^3$Meta AI\\
\small\texttt{shaib.c@northeastern.edu}
}

\begin{document}

\ifcolmsubmission
\linenumbers
\fi

\maketitle

\begin{abstract}
AI ``slop'' is an increasingly popular term used to describe low-quality AI-generated text, but there is currently no agreed upon definition of this term nor a means to measure its occurrence. 
In this work, we develop a taxonomy of ``slop'' through interviews with experts in NLP, writing, and philosophy, and propose a set of interpretable dimensions for its assessment in text. 
Through span-level annotation, we find that binary ``slop'' judgments are (somewhat) subjective, but such determinations nonetheless correlate with latent dimensions such as coherence and relevance. 
Our framework can be used to examine AI-generated text in both detection and binary preference tasks, potentially offering new insights into the linguistic and stylistic factors that contribute to quality judgments. We highlight that fully automated and scalable methods remain an open challenge.
\end{abstract}

\section{Introduction}
``Slop'' has emerged as a term describing generic, low-quality content that appears to have been generated by AI.\footnote{Slop was on the shortlist of Oxford Dictionary's Word of the Year 2024, which claims a ``332\% increase'' in usage: \url{https://corp.oup.com/word-of-the-year/\#shortlist-2024}\label{slop_def}}
Recent news articles offer salient examples of such AI ``slop'', ranging from non-factual claims (``... add nontoxic glue to make cheese stick to a pizza'', ``geologists advise eating at least one rock a day''; \citealp{hoffman2024first}, \citealp{wallace2024}) 
to useless information (``fodder for websites whose only purpose appears to be optimising for [search engines]''; \citealp{mahdawi2025ai}).
Conversations on social media highlight indicators of ``slop'' in LLM text, including overuse of certain terms, low information density, and structural quirks such as lists-as-responses.\footnote{\url{https://x.com/aidan_mclau/status/1884770586276381179}}
Despite the sudden ubiquity of the term, there is no clear definition that enables direct measurement of, nor describes precisely the features of ``slop'' in text.
This gap matters: large-scale surveys, such as Microsoft's Occupational Implications of Generative AI \citep{tomlinson2025working} and Anthropic's Economic Index \citep{handa2025economic} reveal AI is primarily used in writing and information gathering tasks. Defining and measuring ``slop'' may help characterize 
LLM writing. 
Some individuals deeply familiar with AI generated content can reliably detect AI writing on the basis of structural and lexical quirks, even without training \citep{chakrabarty2024art, Russell2025PeopleWF}. 
Yet text can be perceived as ``slop'' even when not generated by AI, and not all AI-generated text reads as ``slop''. 

Our primary aim in this work is to characterize qualities of texts that contribute to their categorization as ``slop.'' Such factors may explain instances where humans mistakenly characterize human-written text as AI-generated, and ``slop'' might provide an explainable metric that accounts for binary preferences between texts collected from human annotators.  We apply principles from measurement theory to conceptualize and operationalize a definition of ``slop'' \citep{bandalos2018measurement, wallach2025position, adcock2001measurement}.  We aim to provide language for articulating style and components of bothersome LLM-generated text, while also providing a framework for measuring such aspects. 

Our main contributions are as follows: 
We first \textbf{introduce a working definition and taxonomy of ``slop''} and map each dimension to automatic metrics where possible (\S\ref{sec:defn}).
To validate this framework, we collect \textbf{span-level annotations from expert writers over 150 news articles and 100 question-answering passages} to provide a fine-grained analysis of slop indicators (\S\ref{sec:measuring}).
Although binary assessments of ``slop'' vary across individuals, we show that \textbf{our taxonomy helps explain which latent qualities (e.g., coherence, relevance, structural features) contribute most to these judgments} (\S\ref{sec:results}). 
We also find that the strength of latent qualities vary based on domain, and that \textbf{our taxonomy provides a useful framework for guiding quality assessment under different tasks} (\S\ref{sec:results}).
For axes that lack adequate automatic measurements (e.g., relevance, coherence, fluency), we find that \textbf{standard text metrics fail to capture annotator preferences, but that syntactic features can recover some performance.}
Finally, we show that capable reasoning \textbf{LLMs also fail to reliably extract and identify ``slop'' in text} (\S\ref{sec:auto_slop}).

\begin{figure}[]
    \centering
    \includegraphics[width=0.9\linewidth]{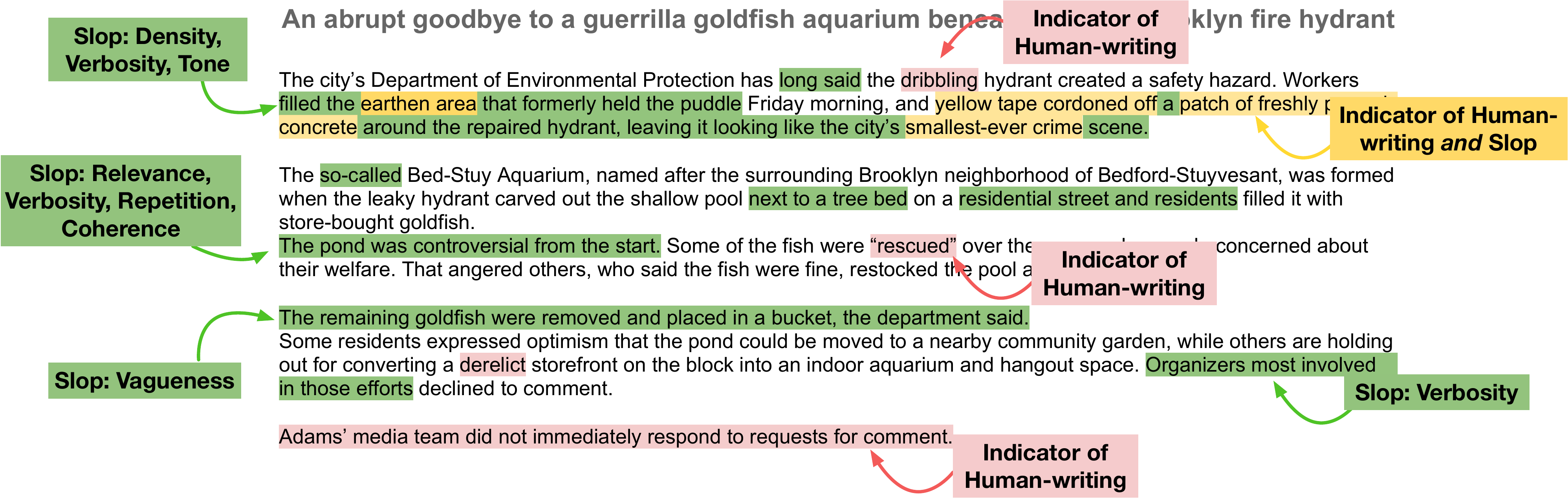}
    \caption{Sample of annotations over a \textit{human-written} news article highlighting indicators of ``slop'' (red; from \citealt{Russell2025PeopleWF}), human-writing (green; ours), and both (yellow). ``Slop'' labels are notably different than indicators of human-written text.}
    \label{fig:slop-fig}
\end{figure} 


\section{Related Work}
\textbf{AI-Text Detection.} There is now a small body of work on discriminating between human- and AI-written texts, e.g., DetectGPT \citep{mitchell2023detectgpt} and Binoculars \citep{Hans2024SpottingLW} provide scores for the likelihood that they were AI generated, and report high discriminant performance (0.95 AUROC). 
\citet{Russell2025PeopleWF} provide an interpretable guide listing key indicators of AI-written text. 
While related, recognizing ``slop'' differs from AI-text detection \emph{in general}, and can be applied to any text source (whether AI-written or not). 
In this work our taxonomy and annotations diverge from those used for AI-text detection in general.

\textbf{Text Diversity.} Prior work has sought to characterize aspects of texts related to how repetitive and \emph{templated} they are. 
\citet{salkar-etal-2022-self}  investigated repeated $n$-grams in LLM outputs in the context of summarization. \citet{shaib-etal-2024-detection} found that modern LLMs are prone to repeatedly generate favoured \emph{syntactic templates}, i.e., sequences of Part-of-Speech (PoS) tags. 
\citet{padmakumar2024does} and \citet{tevet-berant-2021-evaluating} examined lexical and semantic diversity in generated texts, introducing metrics to quantify variation across outputs and emphasizing its importance to generation quality.
These existing efforts have informed the way in which we are thinking about what characterizes writing style and AI ``slop'' and provides automatic measurements for key aspects of ``slop.''

\textbf{Text Quality Measurements.} Text quality has typically been measured using simple surface-level metrics like BLEU \citep{papineni2002bleu} and ROUGE \citep{lin2004rouge}, which can be effective when reference outputs are available (e.g., machine translation evaluation).  
More recent work has recognized that text quality is not monolithic but rather comprises multiple, sometimes competing dimensions that must be measured independently, and accordingly focused on multidimensional frameworks assessing properties of texts. \citet{10.1145/3706598.3713559} provide an editing taxonomy to correct recurring AI-writing flaws such as clichés and unnecessary exposition. Similarly, \citet{bharadwaj2025flattery} show that reward models over-weight 5 superficial writing cues including length, structure, jargon, sycophancy, and vagueness. 
Our work is complementary to measuring quality in general: We  target stylistic patterns unique to LLM writing that are not covered by other taxonomies. 

\textbf{Stylometry.} Stylometry comprises a set of methods that measure surface-level features of writing style, and has been used in the past commonly for LLM-generated and human-written text attribution \citep{stamatatos2009survey, 10.1145/3715073.3715076}. Statistical and computational methods extract features such as relative word, character, and punctuation frequencies (e.g., \citealp{10.1093/llc/17.3.267}; \citealp{doi:10.1126/science.ns-9.214S.237}; \citealp{kevselj2003n}), function word and discourse marker frequencies \citep{segarra2015authorship}, and syntactic structure \citep{zhang2014authorship} to characterize style. Models built over these features can be used to detect plagiarism and authorship attribution. 
These tasks are closely related as we aim to establish a writing style and profile of ``slop'' in this work, and use similar measures for our composite, automatic metrics. 

\section{Defining LLM ``Slop''}
\label{sec:defn}
The Oxford Dictionary defines ``slop'' as: 
    \textit{``[...] material produced using a large language model (LLM), which is often viewed as being \textbf{low-quality or inaccurate.} This type of low-quality, AI-generated material is becoming increasingly visible to people [...], who often view it as \textbf{unwanted or inferior}.''\footref{slop_def}} 
``Slop'' as a construct does not immediately permit measurement: It is difficult to quantify ``low-quality'' or ``unwanted'' text\footnote{Efforts exist (e.g., \citealt{bai2022traininghelpfulharmlessassistant}) to quantify and remove toxic or overtly harmful generations, however our focus is on measuring more subtle forms of ``unwanted'' text.}. 
We propose a composite measure over observable characteristics of text, where we elicit salient characteristics from a set of individuals with a range of relevant expertise.  
Human writing can also read as ``slop'',  but we adopt the above definition and focus on (seemingly) LLM-generated texts.\footnote{Interestingly, what is considered ``slop'' in human-written text can differ in characteristics, and may even serve as an intermediary in writing processes (see Appendix~\ref{app:survey-findings})} 

\begin{wraptable}{r}{0.54\linewidth}
\centering
\resizebox{\linewidth}{!}{%
\begin{tabular}{@{}lllc@{}}
\toprule
\textbf{Themes} & \textbf{Final Codes}   & \textbf{Granular Codes} & \textbf{Count} \\ 
\midrule
\multirow{2}{*}{\textcolor{infoutility}{\textbf{\shortstack[l]{Info. Utility}}}} 
    & Density& IU1: Density            & 5 \\
    & Relevance            & IU2: Relevance          & 9 \\ 
\midrule
\multirow{2}{*}{\textcolor{infoquality}{\textbf{\shortstack[l]{Info. Quality}}}} 
    & Factuality           & IQ1: Factuality         & 7 \\
    & Bias                 & IQ2: Bias               & 2 \\ 
\midrule
\multirow{7}{*}{\textcolor{stylequality}{\textbf{\shortstack[l]{Style Quality}}}} 
    & \multirow{2}{*}{\shortstack[l]{Structure}}& SQ1: Repetition        & 7 \\
    &                      & SQ2: Templatedness     & 2 \\\cmidrule(lr){2-4}
    &                      Coherence& SQ3: Coherence         & 6 \\  \cmidrule(lr){2-4}
    & \multirow{4}{*}{\shortstack[l]{(Aspects of) \\Tone}}& SQ4: Fluency           & 4 \\

    &                      & SQ5: Verbosity         & 5 \\
    &                      & SQ6: Word Complexity   & 1 \\
    &                      & SQ7: Tone              & 3 \\ 
\bottomrule
\end{tabular}%
}
\caption{Themes and codes for slop, count of expert responses containing each.}
 \vspace{-10pt}
\label{tab:taxonomy}
\end{wraptable}

We first solicited detailed definitions of ``slop'' from 19 individuals with a range of expertise across relevant disciplines including writing, journalism, linguistics, NLP, and philosophy (App. Table~\ref{tab:field-counts}). 
This group included PhD students, professors, and industry professionals from the listed disciplines. 
All but one respondent had 3 or more years experience in their field at the time of their response (App. Figure~\ref{fig:yoe}). 
We asked individuals to describe their familiarity with the term ``slop'' in the context of AI-generated content, as well as a description of typical use (if any) of LLMs in their work.
11 experts (58\%) had encountered the term ``slop'' as relates to AI-generated content. 
Most reported using LLMs more than 2 times a week ($n=14$).  The rest mostly used them sporadically ($n=4$) with 1 expert never using them. 
We asked experts to provide a definition and list key characteristics of text that make it ``slop.''
We provide the full survey sent to experts in Appendix~\ref{app:defn}.

Using qualitative content analysis and inductive coding, we map expert definitions of ``slop'' to measurable concepts \citep{Mayring_2000}. 
We begin by identifying key terms in survey responses and building a code list until saturation (i.e., until no new codes are created). 
We then map each response on to one of the following codes: Factuality, Information Density, Bias, Relevance, Repetition, Templatedness, Verbosity, Word Complexity, Tone, Coherence, Fluency, Diversity, Engagement, Vagueness, and Utility. 
The codebook was iteratively updated throughout this process. 
We collapsed redundant codes (e.g., Vagueness and Information Density). 
We also categorize codes into overarching  \textit{themes}: \textcolor{infoutility}{Information Utility}, \textcolor{infoquality}{Information Quality}, and \textcolor{stylequality}{Style Quality}. 
Table~\ref{tab:taxonomy} describes the full code hierarchy within each theme, and the count of responses containing each code tag. 

\subsection{Dataset and Text Generation Details}
\label{app:data}
 
\label{sec:datasets}

We select two datasets to annotate for ``slop'', motivated by two practical observations: First, that LLM-written text is becoming commonplace in reporting news on the internet \citep{tomlinson2025working, Chatterji2025HowPU}. The second is motivated by the use-cases outlined by our experts in \S\ref{sec:defn}, where a majority ($n=9$) reported using LLMs for question answering tasks (Appendix~\ref{app:survey-findings}). 

\textbf{News Article Generation.}
We evaluate ``slop'' over 150 news articles released by \citet{Russell2025PeopleWF}, in which annotators are asked to label texts as being AI-written or not. 
Each unique article has a human-written source and a corresponding AI-written article, generated by either Claude \citep{anthropic2024model}, GPT-4o or o1-pro \citep{jaech2024openai}. 
Additionally, each article includes a ``humanized'' article, where the above models are prompted to avoid obvious LLM-writing indicators. 

\textbf{Retrieval-Augmented QA.}
MS MARCO \citep{Campos2016MSMA} is a large-scale machine-reading-comprehension benchmark comprising real Bing search queries. Each example contains an anonymized user query, a set of candidate web passages retrieved by the search engine, and a human-written answer. We randomly sample 100 queries and generate responses from  Llama-4 Scout\footnote{\url{https://www.llama.com/docs/model-cards-and-prompt-formats/llama4/}}, OLMo-2-13B Instruct \citep{olmo20252olmo2furious}, Mistral-7B \citep{jiang2023mistral7b}, GPT-4o \citep{openai2024gpt4technicalreport}, and Gemma-2-27B \citep{gemmateam2024gemma2improvingopen} (see Appendix~\ref{app:addl-data-details} for prompt and further details).

\subsection{Pilot ``Slop'' Annotation}
\label{sec:pilot}
As an initial validation of the taxonomy, we hire 5 professional copy-editors from the Upwork platform\footnote{\url{https://www.upwork.com/}} to annotate ``slop'' spans in two datasets: News article generation \citep{Russell2025PeopleWF}; and Retrieval Augmented QA (MS MARCO; \citealp{Campos2016MSMA}).\footnote{Exempt Determination obtained from our institutional IRB. 
See Ethical Considerations for approval details.} 
These datasets span different writing styles, expected passage lengths, and serve different purposes for the reader.
We paid annotators at a rate of \$35-45 USD an hour.
Each article took $\sim$10-15 minutes to annotate, with an average of 871 words per article. 
The MS MARCO dataset took $\sim$4-7 minutes to annotate, with an average of 55 words per passage.

We provided annotators with a guide containing codes for indicators of ``slop'' from our expert definition survey (\S\ref{sec:defn}; Appendix Table~\ref{fig:defn-guide}). 
We asked them to read the text in full, and first answer whether they initially perceive the text as ``slop.'' 
We then had annotators label spans in the text (word-level) that instantiate any of the ``slop'' codes. 
Texts may have span annotations even if not initially deemed by the annotator to be ``slop'' overall. 
We provide the full set of questions given to annotators in Appendix Figure~\ref{fig:interface}.
In sum, each annotator annotated the same 10 articles and 10 passages in the pilot round.

Annotation comprising multi-label, multi-span labelling is a difficult task and requires collaborative stages of task training and alignment among annotators. 
 After independently completing the pilot round, annotators met to review codes and annotation strategy. 
Here, the guide was discussed in detail: clarifications around labelling strategy (e.g., whether to select only the most salient codes versus coding every feature), and terminology (e.g., Fluency as a measure of \textit{language naturalness} versus correct grammar) were adjudicated. Most disagreements came from labelling strategy and terminology, rather than disagreements over the text spans.
Table~\ref{tab:taxonomy} shows the final themes and codes after annotator discussion. 
\subsection{Finalized ``Slop'' Taxonomy}
Here, we describe each theme and code after annotator adjudication (See Appendix~\ref{app:breakdown} for a description with examples).

\textcolor{infoutility}{\textbf{Information Utility}}
assesses how effectively a text conveys meaningful and contextually appropriate information. It comprises two key indicators: (i) \textbf{Density}, or the amount of substantive content relative to the length of the text, measured through information-theoretic token entropy \citep{Meister2021RevisitingTU} and propositional idea density \citep{brown2008automatic}, and (ii) \textbf{Relevance}, the alignment of content with task or prompt, measured through expert human annotations \citep{clarke2024llm}.

\textcolor{infoquality}{\textbf{Information Quality}}
describes the accuracy and subjectivity of the presented information. \textbf{Factuality} assesses inaccuracies, hallucinations, or fallacious claims within the text, which require human annotations due to the complexity of automated factual evaluations in the absence of reference texts \citep{ramprasad2025do, laskar2023large}. \textbf{Bias} (Subjectivity) assesses the presence or absence of a necessary subjective or rhetorical perspective, measured by the proportion of subjective words through an established lexicon \citep{wiebe2004learning}.

\textcolor{stylequality}{\textbf{Style Quality}}
addresses properties related to expression and readability. Repetition, identified by lexical repetition metrics \citep{shaib2024standardizingg} and Templatedness, measured via syntactic  structures \citep{shaib-etal-2024-detection} are key features of text \textbf{Structure}. \textbf{Coherence}, defined as poor sentence structure or organization, is evaluated mostly via expert annotations due to the absence of reliable automatic measurements \citep{li2024llms, murugadoss2024evaluatingevaluatormeasuringllms}. Aspects of \textbf{Tone} evaluate the appropriateness and character of language relative to context, and  include issues like excessive formality \citep{fanous2025syceval, yang2024trustllmsmitigateoverconfidence}. We include indicators such as Fluency (\textit{naturalness} of language); Verbosity (passage and sentence length) \citep{zhang2024verbosityneqveracitydemystify}; and Word Complexity, i.e., use of unnecessarily complex vocabulary, measured by Gunning-Fog Index \citep{gunning1952technique} and Flesch-Kincaid Grade Level \citep{kincaid1975derivation}.

\section{Annotating for ``Slop'' in Text}
\label{sec:measuring}
\label{sec:human-annotations}
\begin{wrapfigure}{r}{0.5\textwidth}
    \centering
    \vspace{-2em}
    \includegraphics[width=\linewidth]{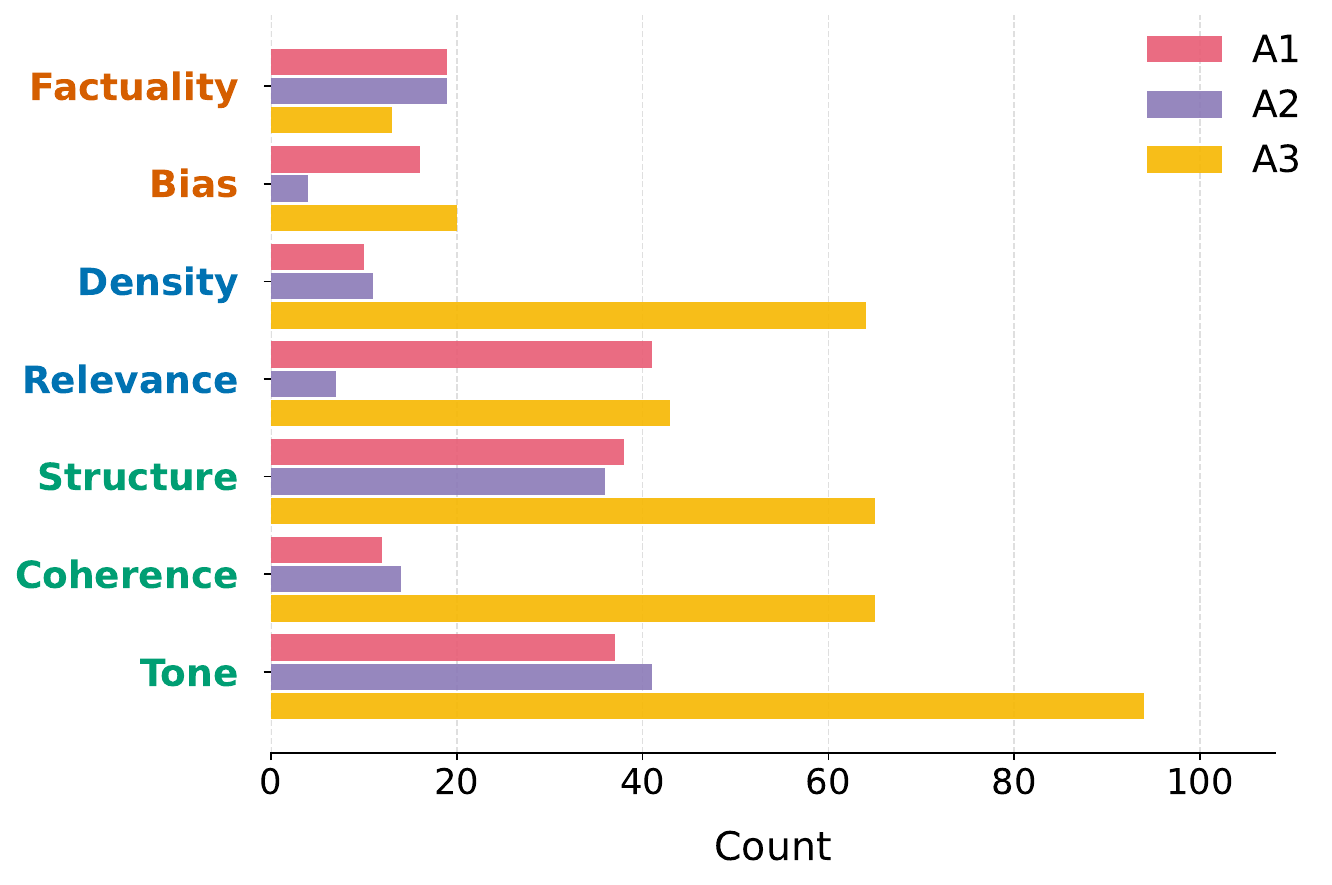}
    \caption{Label counts and distributions for each annotator. Annotators used all codes, but there are individual differences in the code frequencies assigned in each theme. Detailed span statistics in Appendix~\ref{app:span-stats}.}
    \label{fig:radar-chart}
    \vspace{-1.5em}
\end{wrapfigure}
\textbf{Annotation.} We select 3 annotators from our pilot study based on annotation quality and availability for the remainder of the datasets.\footnote{Annotator background and details in Appendix~\ref{app:annotator_bg}} 
Each annotator reviewed 71 news articles (\citealt{Russell2025PeopleWF}; total of 213 articles annotated), and 41 QA passages (MS MARCO, \citealt{Campos2016MSMA}; total of 123 passages). 
We assign a subset of the same 45 news articles and 10 QA passages to all annotators for agreement assessment.

Measuring ``slop'' is difficult: Text can be assigned multiple labels, where a subset represent latent text features (Relevance, Bias, Coherence, Fluency, Tone), compared to directly measurable labels such as Verbosity and Repetition. 
\citet{marchal-etal-2022-establishing} show that the expected overlap between any two annotators in a multi-label setting drops sharply as the number of labels and proportion of double-coded items increases, even after chance-correction. 
This agreement drops further still when labelling latent text features (e.g., coherence) that rely on annotators' mental models of these constructs.

We follow prior work and report a set of appropriate agreement measures for annotations \citep{marchal-etal-2022-establishing}. We report the agreement scores in aggregate over news and QA, and provide the disaggregated values for each domain in Appendix~\ref{app:breakdown}.

\textbf{Span Agreement.} We use the span-level precision measure described in \citet{10.1145/3706598.3713559} to assess if annotators highlighted similar text (Algorithm~\ref{algo_1}).
Span-level precision measures the proportion of highlighted spans that overlap with another annotator’s spans, for each annotator pair.

Here, we consider the highlighted boundaries for precision calculations, which are provided at the character-level by the labelling platform.
At the paragraph level for articles annotators have a pairwise span-level precision of 0.80 (A1--A2), 0.64 (A1--A3), and 0.67 (A2--A3), indicating moderate to high agreement on problematic spans of text, regardless of the assigned code.

\textbf{Label Agreement.} We compute both Cohen's $\kappa$ and Gwet's AC$_1$ over the binary slop label, which indicates whether annotators agree on which documents are overall ``slop.''
Annotator responses had a Cohen's $\kappa$ of -0.12 (A1--A2), 0.11 (A1--A3), and 0.01 (A2--A3), indicating poor agreement. 
Reporting $\kappa$ is consistent with prior work in NLP, but we caution that these scores appear poor due to the low prevalence of ``slop'': Annotators assigned a positive label of ``slop'' to an 36\% of articles on average. 
Gwet's AC$_1$ yields pairwise scores of 0.07 (A1--A2), 0.21 (A1--A3), and 0.16 (A2--A3), indicating poor to fair agreement when correcting for prevalence.

For the binary label agreement, we do not necessarily expect strong agreement here.
We ask annotators to assess ``slop'' labels \textit{before} annotation, and posit that these overall assessments involve a degree of subjectivity. 
Interestingly, annotators show a \textit{fair to moderate agreement} when only considering the news domain, again highlighting the differences in domain and text purpose when deciding on overall ``slop'' assessments (Appendix~\ref{app:breakdown}). 

\begin{wraptable}{r}{0.6\textwidth}
\vspace{-0.5em}
\centering
\resizebox{\linewidth}{!}{%
\begin{tabular}{lllccc}
\toprule
\textbf{Themes} &\textbf{Final Codes}    &\textbf{$\alpha_{MASI}$}& \textbf{$\kappa$} & \textbf{AC$_1$} & \textbf{Prev. (\%)}\\ 
\midrule
\multirow{2}{*}{\textcolor{infoutility}{\shortstack[l]{Info. Utility}}} 
    &Density         &\multirow{2}{*}{\shortstack[l]{0.37}}& 0.17&  0.55&49.1\\
    &Relevance             && 0.17&  0.24&69.1\\ 
\midrule
\multirow{2}{*}{\textcolor{infoquality}{\shortstack[l]{Info. Quality}}} 
    &Factuality            &\multirow{2}{*}{\shortstack[l]{0.44}}& 0.18&  0.75&30.9\\
    &Bias                  && 0.15&  0.73&32.7\\ 
\midrule
\multirow{3}{*}{\textcolor{stylequality}{\shortstack[l]{Style Quality}}} 
    &\multirow{1}{*}{\shortstack[l]{Structure}} 
     &\multirow{3}{*}{\shortstack[l]{0.37}}& 0.13&  0.15&74.5\\
    &Coherence && 0.11&  0.51&50.9\\  
    &\multirow{1}{*}{\shortstack[l]{Tone}}
     && 0.01&  0.24&70.9\\
\bottomrule
\end{tabular}%
}
\caption{Agreement ($\alpha_{MASI}, \kappa, AC_{1}$) and label prevalence (Prev. \%) for ``slop'' codes across all data.}
\label{tab:agreement-all}
\vspace{-0.7em}
\end{wraptable}

\textbf{Taxonomy Agreement.} In the ``slop'' taxonomy labelling task, multiple codes can be assigned to a span, and multiple spans can exist in a document. 
We first aim to understand the convergence of the code sets assigned to each document. 
Following \citet{marchal-etal-2022-establishing}, we calculate Krippendorf's $\alpha_{\text{MASI}}$ which measures set agreement chance‑corrected for partial overlaps. 
Next we try to evaluate the individual reliability of each code. 
We report both Cohen's $\kappa$ (for pairwise), Fleiss  $\kappa$ (for three-way) and Gwet's AC$_1$, noting that the AC$_1$ scores will be a more reliable assessment in this setting as annotators have differing rates of label assignment, shown in Figure~\ref{fig:radar-chart}. 



Table~\ref{tab:agreement-all} (all) and Appendix Table~\ref{tab:agreement-pairwise} (pairwise) report agreements calculated across finalized codes and overall themes. 
After three calibration rounds of guided discussion, we report theme‐level Krippendorff’s $\alpha_{MASI}$ of 0.37 (\textcolor{infoutility}{Info. Utility} and \textcolor{stylequality}{Style Quality}) and 0.44 (\textcolor{infoquality}{Info. Quality}). These values fall within the ``moderate-to-challenging'' band ($\alpha \approx 0.10 - 0.50$) for high-entropy, construct-level annotation \citep{marchal-etal-2022-establishing}, indicating that annotators consistently overlap on at least some taxonomy labels within each theme, but that full label-set consensus is difficult to achieve. 
At the code level, Density (AC$_1 = 0.55$), Factuality (AC$_1 = 0.75$), Bias (AC$_1 = 0.73$), and Coherence (AC$_1 = 0.51$) reach  agreement above the 0.5 reliability threshold, indicating dependable annotation. In contrast, cognitively demanding constructs requiring higher-level, interpretable judgements such as Relevance, Structure, and Tone fall near the ``moderate-to-challenging'' band. This indicates that these labels can be used for research, but with caution.

\section{What is ``Sloppy'' Text?}
\label{sec:results}
\definecolor{darkgrey}{gray}{0.3}
\newmdenv[
  skipabove=3pt,
  skipbelow=3pt,
  leftmargin=2pt,
  rightmargin=2pt,
  innerleftmargin=4pt,
  innerrightmargin=4pt,
  innertopmargin=2pt,
  innerbottommargin=2pt,
  linewidth=0pt,
  fontcolor=darkgrey,
  font=\small\sffamily
]{greybox}

\newcommand{\hliq}[1]{\sethlcolor{infoqualityhl}\hl{#1}}
\newcommand{\hliu}[1]{\sethlcolor{infoutilityhl}\hl{#1}}
\newcommand{\hlsq}[1]{\sethlcolor{stylequalityhl}\hl{#1}}

\definecolor{neutralgreen}{RGB}{34, 139, 34}
\definecolor{lightred}{RGB}{220, 100, 100}

\newcommand{\greyX}{\textcolor{lightred}{\ding{55}}}
\newcommand{\greencheck}{\textcolor{neutralgreen}{\ding{51}}}

We now present results from all annotations collected across the news and QA datasets. 
Our analysis includes both a \textit{combined} setting across all domains, as well as separate evaluations by domain (news vs. QA). 
For each setting, we report results aggregated across all annotators
(See Appendix~\ref{app:breakdown} for individual plots). 
The combined analysis highlight common ``slop'' features shared across all data and annotators, while the disaggregated evaluation show  variations that may arise from annotator subjectivity or domain-specific patterns.

We first construct ``slop'' features as an aggregated count and presence of the span-level codes across annotators. We fit a logistic regression model\footnote{Using {\tt statsmodels}, version 0.14.0.} with these features as the independent variables and a binarized ``slop'' label as the (single) dependent variable.
This allows us to evaluate whether aggregated patterns in the span-level taxonomy are associated with the binary slop judgments. 
Features with adjusted $p<0.05$ (after Bonferroni correction) are considered statistically significant predictors of whether annotators label texts as ``slop.'' Figure~\ref{fig:codes_domain_split} shows that the individual axes influencing ``slop'' assessments vary slightly in the domain-specific regressions.

\subsection{Results}
\label{sec:results_combined}

\begin{figure*}[]  
  \centering
  \begin{minipage}[t]{0.375\textwidth}
    \centering
    \includegraphics[width=\linewidth]{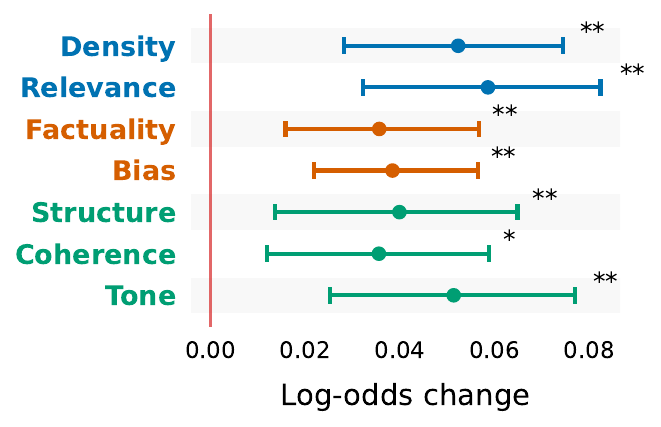}
    \label{fig:slop-all-codes}
  \end{minipage}
  \hspace*{\fill}
  \begin{minipage}[t]{0.28\textwidth}
    \centering
    \includegraphics[width=\linewidth]{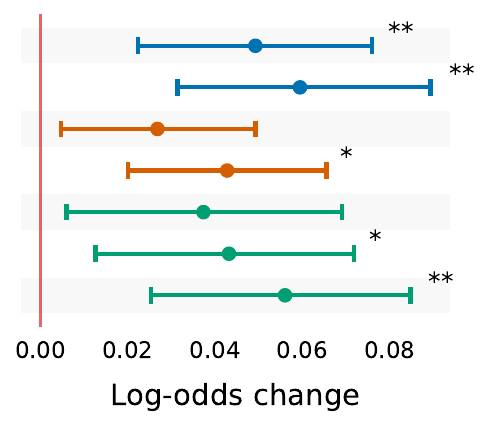}
    \label{fig:news_codes}
  \end{minipage}
  \hspace*{\fill}
  \begin{minipage}[t]{0.30\textwidth}
    \centering
    \includegraphics[width=\linewidth]{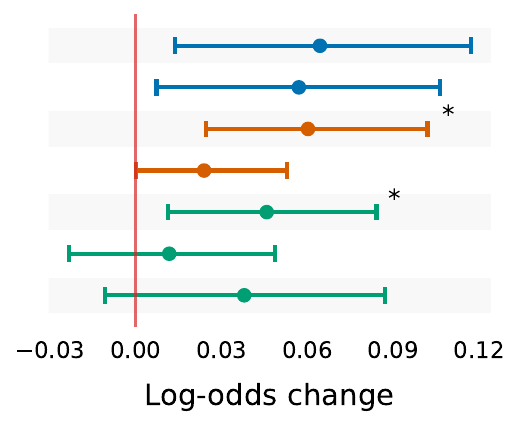}
    \label{fig:ms_marco}
  \end{minipage}

  \caption{``Slop'' codes most predictive of the overall positive label for (left) the entire corpus, (middle) news, and (right) MS MARCO. * $p < 0.05$, and ** $p < 0.01$. }
  \label{fig:codes_domain_split}
\end{figure*}

We first confirm that more annotated spans in  documents does in fact correlate with assessments of ``slop'' across annotators: $\rho_{news} = 0.70, \rho_{ms\_marco} = 0.51, \rho_{all} = 0.63$. 
Across the annotations, all seven codes are significant (positive) predictors of an item being labelled ``slop,'' which empirically validates the taxonomy built in~\S\ref{sec:defn}.
The strongest predictors are text issues like \textbf{Relevance} ($\hat{\beta}=0.06$), \textbf{Density} ($\hat{\beta}=0.05$), and \textbf{Tone} ($\hat{\beta}=0.05$).
The combined analysis of ``slop'' codes shows that broadly the quality deficit in the text is significant across all style, information quality, and utility themes (Fig.~\ref{fig:codes_domain_split}, left). Text lacking relevance and information, or containing factual errors or biased language, is consistently labelled as ``slop'' across domains. 

\begin{table}[]
\centering
\small
\resizebox{0.96\linewidth}{!}{
\begin{tabular}{@{}p{2cm}p{11.2cm}@{}}
\toprule
\textbf{Code category} & \textbf{Highlighted span} \\ \midrule 
\textcolor{infoutility}{Relevance} & ``During the Roman Empire, physicians developed techniques to repair injured gladiators and soldiers, including methods for treating facial injuries and performing basic skin grafts. \hliu{The field experienced a significant evolution during the Renaissance, as European surgeons began documenting and sharing their techniques more widely.}'' \\
\textcolor{infoquality}{Factuality} & ``[...] leading to more frequent and severe heatwaves. \hliq{``Climate change is like adding steroids to our weather," says Dr. Michael Oppenheimer, a climate scientist at Princeton.}''  \\
\textcolor{stylequality}{Structure} & ``But did you know there's another important number-sort of like a \hlsq{"secret" code---printed just beneath the sell-by date}? [...] Find \hl{the secret code, which is usually near the sell-by date.}'' \\
\textcolor{stylequality}{Tone} & ``The very power of the word [``witch"] lies in its imprecision. \hlsq{It is not merely a word but an archetype, a cluster of powerful images." The uncertainty of exactly what a witch is forms part of the titillation}'' \\
 \bottomrule
\end{tabular}
}
\caption{Text marked by all annotators. \textcolor{infoutility}{Relevance}: all marked as irrelevant. \textcolor{infoquality}{Factuality}: Inaccurate quote attribution. \textcolor{stylequality}{Structure}: marked for repeated text. \textcolor{stylequality}{Tone}: marked for coherence, fluency.}
\label{tab:issue-examples}
\end{table}

\textbf{News.} For news articles, issues with style quality (Coherence, Tone), information utility (Density, Relevance), and Bias are significant predictors. Annotators find text that is verbose, off-topic, or that contains tonal/framing issues in news articles as indication of ``slop'' (Fig.~\ref{fig:codes_domain_split}, middle).

\textbf{MS MARCO.} For QA tasks, factuality and structural issues are the strongest predictors of ``slop'' for all annotators. 
Text from MS MARCO passages are short, which may reduce the  significance of the Density, Relevance, and tonal codes. Here, annotators deem answers that are ill-structured and  factually incorrect as being indicative of ``slop''  (Fig.~\ref{fig:codes_domain_split}, right). 

The disaggregated analysis shows that features important for ``slop'' vary based on the purpose of the text: Factual and structural issues are more significant for QA data, while stylistic and utility issues are more prominent for news articles. \textbf{This distinction indicates the importance of evaluating LLM-written texts with respect to domain to contextualize their quality.}
\footnote{We provide further qualitative assessment of the label distribution across topics and sources within the news domain in Appendix~\ref{app:qualitative_domain_assessment}.}

\section{Automatically Measuring Slop}
\label{sec:auto_slop}
Building on the annotation analysis, we investigate whether ``slop'' can be measured with automatic methods. We evaluate standard text metrics and LLM-based approaches for capturing both the underlying codes reflected in the annotations and the overall binary ``slop'' assessments. Appendix~\ref{app:wqrm} provides an additional comparison to existing writing quality rewards models.  

\subsection{Lexical, Syntactic, and Stylometric Features}
\label{sec:auto-linear}
Table~\ref{tab:taxonomy-auto} provides the entire slop code taxonomy and a mapping to existing automatic text evaluation metrics. 3 out of 5 codes that were significant features of slop assessments in \S\ref{sec:results} do not have reliable metrics, motivating the need for human annotations. 
Nonetheless, we examine 
all available automatic metrics to assess their ability to capture the latent qualities of text in our taxonomy.
We also include stylometric features: word-length frequencies \citep{doi:10.1126/science.ns-9.214S.237}, stop word frequencies; hapax legomena \citep{garcia2006function},  and syntactic features such as tree depth measures, branching factors, 
dependencies \citep{hollingsworth2012using}, passive voice markers, noun- and verb-phrase counts \citep{zhang2014authorship, stamatatos2009survey, grieve2007quantitative}.

We examine linear models: logistic regression, and a pipeline approach following \cite{zhang2014authorship} where we first apply principle component analysis (PCA) then linear discriminant analysis (LDA). 
We test the following feature sets: syntax-only, lexical-only, automatic features (Table~\ref{tab:taxonomy-auto}, Appendix Table~\ref{tab:full-feature-list}), and all features combined. 
Many automatic text measures have high overlap in information (e.g., \citealp{shaib2024standardizingg}, and Figure~\ref{fig:correlation}), which can lead to multicollinearity issues in regression models. 
To address this and handle class imbalance, we use $\ell$2 regularization with $\alpha=1.0$ and class weighting. 
We standardize all predictors and drop highly correlated features with threshold $\geq 0.95$.

\begin{figure*}[]
    \centering
    \resizebox{0.9\textwidth}{!}{%
    \begin{subfigure}[t]{0.48\textwidth}
        \centering
        \includegraphics[width=\linewidth]{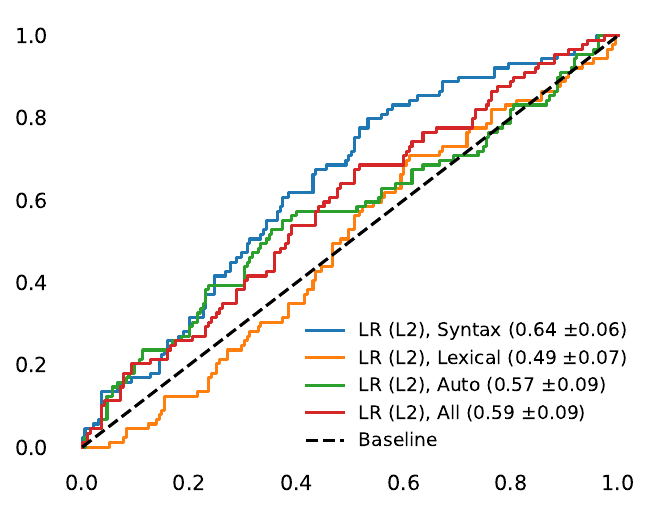}
        \label{fig:combined_auroc_LR}
        \vspace{-1em}
    \end{subfigure}
    \hfill
    \begin{subfigure}[t]{0.48\textwidth}
        \centering
        \includegraphics[width=\linewidth]{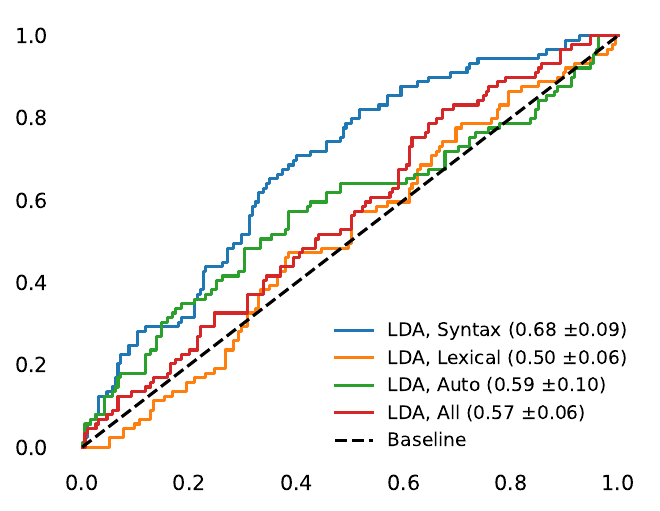}
        \label{fig:combined_auroc_LDA}
        \vspace{-1em}
    \end{subfigure}
    }%
    \caption{AUROC predicting the binary ``slop'' label across both the news and QA domains using (left) a logisitic regression model and (right) the PCA+LDA model. The syntax-only feature set performs better than lexical or automatic features.}
    \vspace{-1em}
    \label{fig:auroc_combined}
\end{figure*}

\textbf{Results.} Figure~\ref{fig:auroc_combined} shows the AUROC for predicting the binary slop label under the (a) logistic regression model and the (b) LDA model. Surprisingly, using only syntax-derived features achieves the best performance on this task, under both the linear model (AUROC 0.64, $\pm0.06$) and the PCA+LDA approach (AUROC 0.68, $\pm0.09$). 
Cognitive studies have shown that humans can struggle to consciously or explicitly identify features of syntax \citep{batterink2013human}, 
and we posit that this may explain the gap between our taxonomy and 
practical automatic feature selection.
We leave studies examining the full range of syntactic features as ``slop'' style predictors for future work. 

\subsection{Can LLMs Self Identify ``Slop''?}
\label{sec:llm-spot-slop}
Recent text evaluations have prompted LLMs to judge text qualities not readily captured by existing metrics (e.g., \citealt{liu2023g, zheng2023judging}). 
This is usually done zero-shot, providing instructions for evaluation. 
Here we test three LLMs (GPT-5, Deepseek-V3, and o3-mini) for their ability to (a) predict binary ``slop'' labels and (b) extracting ``slop'' spans. In both cases, we provide the full ``slop'' guide given to annotators (see Appendix~\ref{app:llm-evals} for prompting details). Finally, we also test a per-code prompting setting, where we issue one GPT-5 call per code to independently extract spans, rather than providing the full guide in a single pass. 

\textbf{Results: Classifying Slop} Table~\ref{tab:combined-assessment} (left) shows the results of binary slop prediction for GPT-5, Deepseek-V3, and o3-mini. Agreement with human annotators is low   ($\kappa$ $\approx$ 0). 
Models under-predict the slop label (0.03-0.08), especially relative to humans (0.35). 
Both recall (0.08-0.12) and precision (0.25-0.42) are (very) low across all models, showing LLMs do a poor job at this task. 


\begin{table*}[]
\centering
\small
\begin{minipage}[t]{0.48\textwidth}
\centering
\begin{tabular}{lccc}
\toprule
\textbf{Model} & \textbf{$\kappa$} & \textbf{Pr./R} & \textbf{Pct. Slop}\\
\midrule
GPT-5 & 0.01 & 0.38/0.12 & 0.08\\
Deepseek-V3 & -0.01 & 0.25/0.08 & 0.03\\
o3-mini & 0.03 & 0.42/0.08 & 0.07\\ 

\midrule
Human & -- & -- & \textbf{0.34} \\
\bottomrule
\end{tabular}
\label{tab:binary-assessment}
\end{minipage}
\hfill
\begin{minipage}[t]{0.48\textwidth}
\centering
\begin{tabular}{lcll}
\toprule
\textbf{Model} & \textbf{k} & \textbf{Precision} & \textbf{Recall}\\
\midrule
GPT-5 & 0 & 0.14 & 0.11\\
& 1 & 0.14 & 0.11\\
& 3 & 0.16 & 0.13\\
& 5 & 0.13 & 0.10\\
\midrule
Qwen-7B* & --- & \textbf{0.32} & \textbf{0.30}\\
\bottomrule
\end{tabular}
\label{tab:span-extraction}
\end{minipage}
\caption{(left) Binary slop assessment: 0-shot prompting relative to human labels (humans assigned positive slop label to ~35\% of data). (right) Span-level extraction: Character-level precision and recall in zero- and few-shot settings. *Qwen-7B is fine-tuned.}
\label{tab:combined-assessment}
\end{table*}



\textbf{Results: Extracting ``Slop'' Spans}
On average, GPT-5 extracted longer text spans than human annotators (mean 58 vs. 41 characters, respectively). However, span-level alignment with human annotations is low. Table~\ref{tab:combined-assessment} (right) shows the results of prompting GPT-5 zero-shot to extract spans, and with in-context examples ($k \in [1, 3, 5]$). We report the \textit{character-level} precision and recall.\footnote{Implementation details in Appendix~\ref{app:pseudocode-precision}.} GPT-5 achieves a precision of 0.08, recall of 0.12. Providing examples up to $k=5$ does not improve precision nor recall by much: reporting precision of 0.13, and recall of 0.19. Further, the additional in-context examples do not significantly change which spans are extracted, there is a consistent F1 overlap of 0.65-0.67 between each $k$ setting and $k=0$. While the higher recall relative to precision suggests the model can identify some relevant text spans, the overall overlap remains low.

\textbf{Results: Single-Code Span Extraction}
GPT-5 produced at least one span in 80\% of articles, compared to 82\% in the human-labelled data (precision 0.81, recall 0.83).
This suggests the model largely agrees about which articles contain ``slop'' spans. 
For span precision/recall, we find similar average performance to the full guide prompting method (cf. Table~\ref{tab:combined-assessment} (right) and Table~\ref{app:pr-single-llj} (left)). 
Codes such as Structure and Density show higher span precision, suggesting that they may be easier to identify. While separate calls increase the number of highlighted spans, the predicted spans are on average much shorter than human annotations (mean $\approx81$ vs. 163 characters), and the model produces more spans in total (374 predicted vs. 135 human-annotated). 
We provide additional single-code details and an analysis of model reasoning outputs in Appendix Tables~\ref{app:span-length-single-code},~\ref{app:single-code-qual}.

\begin{table}
\centering
\begin{minipage}[t]{0.47\textwidth}
\centering
\setlength{\tabcolsep}{4.5pt}
\begin{tabular}{lccc}
\toprule
Code & Precision & Recall & Presence F1 \\
\midrule
Bias        & 0.15 & 0.03 & 0.84 \\
Coherence   & 0.11 & 0.02 & 0.83 \\
Density     & 0.26 & 0.08 & 0.90 \\
Factuality  & 0.12 & 0.02 & 0.82 \\
Relevance   & 0.05 & 0.02 & 0.84 \\
Structure   & 0.25 & 0.03 & 0.82 \\
Tone        & 0.15 & 0.03 & 0.80 \\
\midrule
Mean        & 0.16 & 0.03 & 0.84 \\
\bottomrule
\end{tabular}
\end{minipage}
\hfill
\begin{minipage}[t]{0.44\textwidth}
\centering
\begin{tabular}{llr}
\toprule
\multirow{2}{*}{Spans Per Theme} & Human & 1.29 \\
                                 & GPT-5 & 1.53 \\
\cmidrule(lr){1-3}
\multirow{2}{*}{Span Length} & Human & 162.63 \\
                                    & GPT-5 & 80.77 \\
\cmidrule(lr){1-3}
\multirow{2}{*}{Total Spans}     & Human & 135 \\
                                 & GPT-5 & 374 \\
\bottomrule
\end{tabular}
\end{minipage}
\caption{(left) Character precision, recall, and code presence F1 of the extracted ``slop'' spans using GPT-5 over the News dataset (cf. Table~\ref{tab:combined-assessment}). (right) Average span character length and counts comparison to human highlights.}
\label{app:pr-single-llj}
\end{table}

\section{Discussion}
LLMs are often deployed as 
cheap alternatives for human preference judgments in alignment and evaluation \citep{bharadwaj2025flattery}, however
our findings highlight important limitations. 
Unlike reasoning tasks where rewards are verifiable, for subjective tasks there is a significant risk of miscalibration. Prior work has documented these issues:  \cite{chakrabarty2025ai} and \cite{Gooding2025WritingAA}, for example, show that LLMs struggle to select high-quality writing actions as judged by human experts, often treating suboptimal and optimal interventions as equally acceptable. 
This leads to low quality text that is often referred to as ``slop". 

A recent study from OpenAI \cite{Chatterji2025HowPU} shows that almost 50\% of ChatGPT usage focuses on writing (28.1\%) and information seeking (21.3\%) tasks. 
To ensure better alignment in such areas, we present the first 
attempt at qualitatively characterizing ``slop’’ in LLM-generated text. 
Our findings suggest that \textcolor{infoquality}{Information Quality}, \textcolor{infoutility}{Information Utility}, and \textcolor{stylequality}{Style Quality} are important axes of text evaluation. Further, granular codes within each axis can vary in strength based on the domain, or the purpose of the text. 
We show that our taxonomy provides a useful framework for assessing writing across domains, beyond accuracy- or reference-based metrics. 

While overall ``slop'' judgments are somewhat subjective, our analysis shows that an increase in issues along these axes increases the likelihood of a text being judged as ``slop''.  We also show that current evaluation practices are not sufficient for automatically measuring ``slop''. Existing automatic text metrics fail to capture whether generated text is genuinely useful or well-written relative to ``slop''. Neither LLMs-as-judges nor linear models trained over these features are able to fully approximate human assessments of ``slop,'' however we hope the taxonomy can guide future improvements of LLM-based reward models. While our analysis focuses on text written in English, we hope the framework introduced here can support future analyses of other languages.

\section{Ethics Statement}
This study was reviewed and deemed exempt by the Northeastern Review Board (IRB: \#25-03-28). Prior to their involvement in the project, all annotators were briefed on the purpose of the research and provided informed consent (Appendix~\ref{app:consent_sec}). 
We prioritized fair compensation, which annotators set prior to participation on the Upwork platform. 
Our dataset consists of publicly available news and QA passages, and no personally identifiable information was used. The focus of this work is on characterizing properties of AI-generated text; it does not target or analyze individuals but rather professional assessments of writing quality.

\section{Reproducibility Statement} The taxonomy, annotation guidelines, survey protocols, and detailed descriptions of data sources are included in Appendix~\ref{app:defn}, \ref{app:annotation}, and \ref{app:llm-evals}. Code for data processing and ``slop'' assessments from LLMs are provided in the supplementary materials. All experiments are described with hyperparameters and settings in \S\ref{sec:auto_slop} and Appendix~\ref{app:autometrics}. 
We collect annotations on only English texts and acknowledge this is an inherent limitation of the work. While we expect the hallmarks of “slop” to be consistent across languages, future work should empirically verify this. We believe our work provides data and a framework to facilitate comparisons of “slop” evaluations in other languages.
 
\subsubsection*{Acknowledgements}
This work was supported in part by the National Science Foundation (NSF) grant RI-2211954. 

\bibliography{colm2026_conference}
\bibliographystyle{colm2026_conference}

\clearpage
\appendix
\section{LLM Disclosure}
LLMs were used to generate data (Sections~\ref{sec:datasets},~\ref{sec:pilot}) and for self-evaluation over axes of ``slop'' (Section~\ref{sec:llm-spot-slop}). LLMs were also used to generate Matplotlib/Seaborn code for the plots. 

\section{Definition Collection}
\label{app:defn}
We provide the full survey sent to annotators in  Table~\ref{app:survey-questions}. Here, we show the full set of questions and answer options in the survey. Annotators explicitly provided permission at the end to share anonymized and aggregate responses to the survey. 

\begin{table}[h]
\begin{longtable}{p{0.45\textwidth} p{0.45\textwidth}}
\toprule
\textbf{Question} & \textbf{Answer Options} \\
\midrule
\endfirsthead
\toprule
\textbf{Question} & \textbf{Answer Options} \\
\midrule
\endhead
\midrule
\endfoot
\bottomrule
\endlastfoot

What is your primary field of work? & Natural Language Processing \newline Linguistics \newline Writing (e.g., Copywriting, Journalism) \newline Other: [Free text] \\
\midrule

How many years of experience do you have in your field? & 1-2 \newline 3-5 \newline 5-8 \newline 9+ \\
\midrule

Have you encountered the term ``slop'' before, as it relates to generated text or images? & Yes \newline No \\
\midrule

If you answered ``Yes'' to the above question, please describe which contexts you've encountered the term in (e.g., in a news article, in a podcast, in discussions) & Free text \\
\midrule

How often do you use large language models (LLMs) in your work? & Never \newline Rarely (sporadically) \newline Sometimes (2 times a week) \newline Often (3-4+ times a week) \\
\midrule

If you use LLMs once or more a week, please select the type of tasks you use them for. & Ideating \newline Writing \newline Rewriting \newline Summarizing documents/texts \newline Creative writing \newline Question Answering (general) \newline Question Answering (specific, based on an input document) \newline Other: [Free text] \newline \textit{(Check all that apply)} \\
\midrule

Please define ``slop'' as it relates to either AI-generated or human-written text. Please be as specific as possible. Think of the contexts in which you use or encounter LLM-generated texts to help guide your answer. & Free text \\
\midrule

Which key characteristics in text are associated with ``slop''? & Free text \\
\midrule

Is all AI-generated text ``slop''? & Yes \newline No \newline Maybe \\
\midrule

Any other thoughts you would like to share about the definition of ``slop'' or its role in text? & Free text \\ 
\bottomrule
\end{longtable}
\caption{Anonymous survey sent to experts to collect definitions of ``slop.''}
\label{app:survey-questions}
\end{table}

\section{Additional Survey Findings}
\label{app:survey-findings}
\begin{figure}[t]
    \centering
    \begin{subfigure}[b]{0.45\textwidth}
        \centering
        \includegraphics[width=0.8\linewidth]{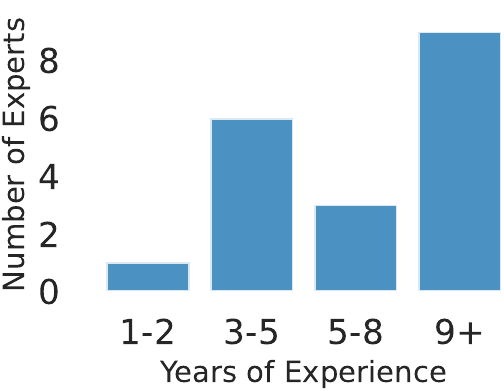}
        \caption{Years of experience for each expert.}
        \label{fig:yoe}
    \end{subfigure}
    \hfill
    \begin{subfigure}[b]{0.43\textwidth}
        \centering
        \resizebox{\linewidth}{!}{%
        \begin{tabular}{lc}
        \toprule
        \textbf{Field of Work} & \textbf{Count} \\
        \midrule
        Natural Language Processing & 6 \\
        Writing & 5 \\
        Engineering & 2 \\
        Machine Learning & 3 \\
        Linguistics & 2 \\
        Philosophy & 1 \\
        \midrule
        \textbf{Total} & 19 \\
        \bottomrule
        \end{tabular}}
        \caption{Work areas of experts surveyed.}
        \label{tab:field-counts}
    \end{subfigure}
    \caption{Expert demographics: (a) distribution of years of experience and (b) fields of expertise.}
    \label{fig:expert-demographics}
\end{figure}

Experts provided their years of experience in their reported fields (Table~\ref{tab:field-counts}), which we report in aggregate in Table~\ref{fig:yoe}. 
Most annotators had $\geq$3 years of experience, and many had professional experience in the field of NLP or writing. 
Additionally, experts identified characteristics of ``slop'' that can appear in human-written text, but all point to qualities that serve a different purpose than those identified for AI-generated text. 

\begin{table}[t]
\centering

\begin{tabular}{p{\textwidth}}
\toprule
\textbf{Expert Response (trunc)} \\
\midrule
``...`slop' text from humans as something that is \textbf{very generic and overly verbose}—perhaps \textbf{excessive marketing copy} or \textbf{rambling prose} that is published when it really should have been heavily edited. I wouldn't call it `slop' if it was simply a first draft—it's the \textbf{brazen publishing} (online or IRL) of content that \textbf{wastes a reader's time} that befits the term best.'' \\
\midrule
``The classic \textbf{highschool transitional words} that students just learning to write essays use, \textbf{without variance}.'' \\
\midrule
``Slop in certain human contexts may perform a \textbf{useful function as an intermediary step} in the author's process (consider certain kinds of \textbf{notes or student writing}, for example).'' \\
\midrule
``...there are many responses from humans that are \textbf{hastily written without critical thoughts}.'' \\
\bottomrule
\end{tabular}
\caption{Sample responses describing how ``slop'' may manifest in human written text.}
\label{app:human-resp}
\end{table}

\section{Consent Forms}
\label{app:consent_sec}
\begin{figure}
    \centering
    \includegraphics[width=0.6\linewidth]{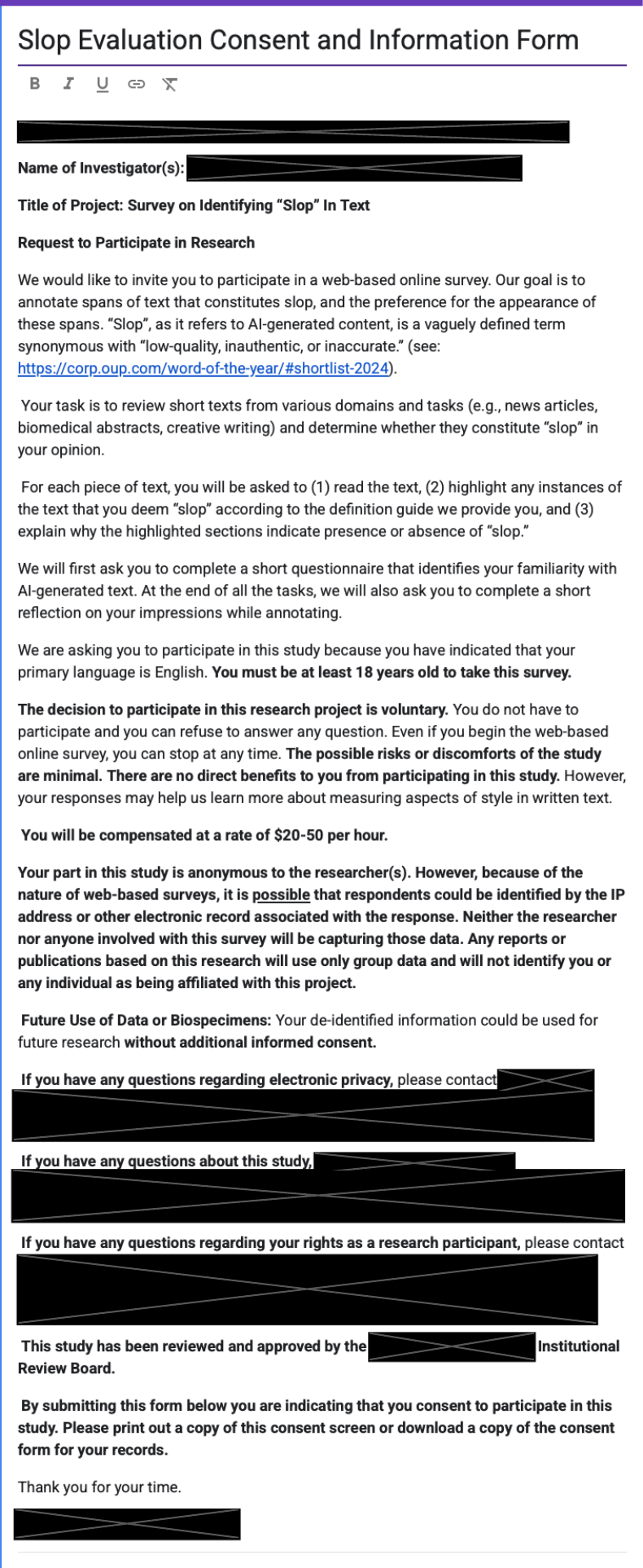}
    \caption{``Slop'' Evaluation consent form given to annotators prior to the study.}
    \label{fig:consent}
\end{figure}

All annotators were briefed on the study and provided explicit consent to share their anonymized responses. We provide the consent form given to participants in Figure~\ref{fig:consent}.

\section{Expert Annotator Details}
\label{app:annotator_bg}

Here, we provide details of the backgrounds and expertise of our selected annotators. 
Our analyses are limited to English texts so we hired annotators who are fluent in English.
All annotators are native English speakers from North America. Each annotator had between 15 and 30 years of experience working as professional copy-editors and writers, in Education, Publication, and Business fields. Annotators have worked additionally as writers for print and online professional publications, and as educators teaching writing.

\section{Annotation Details}
\label{app:annotation}

\begin{figure}[t]
    \centering
    \includegraphics[width=\linewidth]{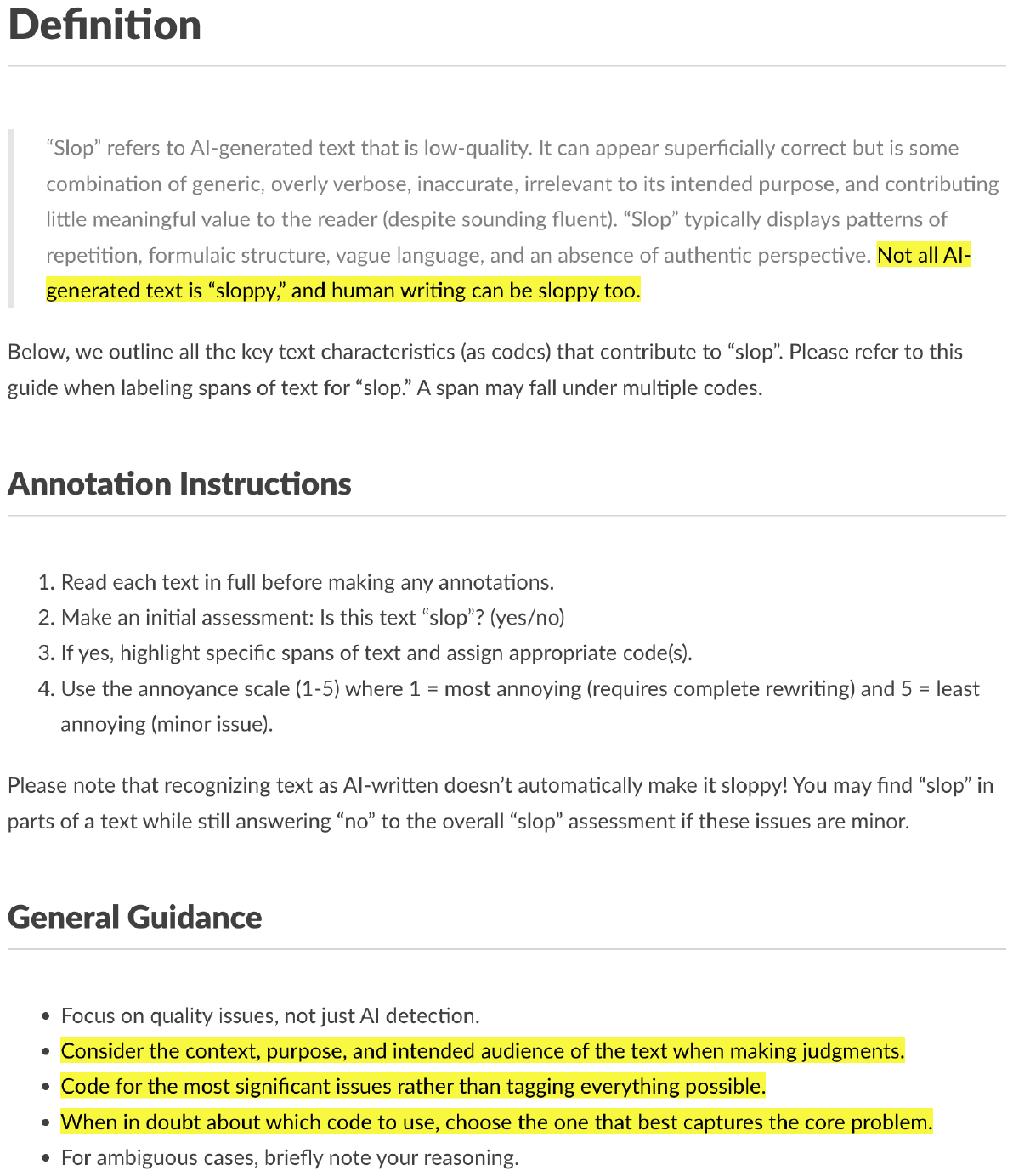}
    \caption{Definition Guide presented to annotators (along with Table~\ref{app:code-defn}) to reference.}
    \label{fig:defn-guide}
\end{figure}

We build our annotation interface using LabelStudio, shown in Figure~\ref{fig:interface}. 
Annotators are instructed to first answer the ``Initial Assessment'' of ``slop.'' After this question is answered, they proceed with the annotation with the codes.
We provide annotators with a PDF document containing the specific definition of slop (Fig.~\ref{fig:defn-guide}). This guide also contains a definition of each ``slop'' code with examples of how each code can appear in text (Table~\ref{app:code-defn}).

\begin{figure}[t]
    \centering
    \includegraphics[width=\linewidth]{images/interface_wide.pdf}
    \caption{questions given to annotators in interface (LabelStudio)}
    \label{fig:interface}
\end{figure}

\begin{longtable}{p{0.8cm} p{1.8cm} p{5cm} p{5cm}}
\caption{``Slop'' Codes with Examples}\label{app:code-defn}\\
\toprule
\textbf{Code} & \textbf{Name} & \textbf{Description} & \textbf{Example} \\
\midrule
\endfirsthead

\toprule
\multicolumn{4}{l}{\tablename~\thetable{} \textit{(continued)}}\\
\midrule
\textbf{Code} & \textbf{Name} & \textbf{Description} & \textbf{Example} \\
\midrule
\endhead

\midrule
\multicolumn{4}{r}{\textit{Continued on next page}}\\
\bottomrule
\endfoot

\bottomrule
\endlastfoot

\multicolumn{4}{l}{\textbf{\textcolor{infoquality}{Information Quality}}} \\
\midrule
IQ1 & Factuality &
\begin{minipage}[t]{\linewidth}\vspace{-6pt}
Incorrect or fabricated information \\ 
Misleading or fallacious claims
\end{minipage} &
``Dr. Sarah Johnson of Harvard University published groundbreaking research on this topic in 2022.'' \newline \textit{(Slop if Dr. Johnson doesn't exist, isn't at Harvard, or didn't publish such research)} \\
\midrule
IQ2 & Bias &
\begin{minipage}[t]{\linewidth}\vspace{-6pt}
Text that feels too ``objective'' when subjectivity is appropriate \\ 
Missing rhetorical point of view when needed \\ 
Lack of appropriate perspective based on context \\ 
Content that seems detached when engagement is required \\ 
The presence of inappropriate perspective or assumptions
\end{minipage} &
``The economic policy changes of 2023 were universally beneficial.'' \newline \textit{(Slop because it presents a one-sided view of complex policy impacts)} \\
\midrule

\multicolumn{4}{l}{\textbf{\textcolor{infoutility}{Information Utility}}} \\
\midrule
IU1 & Information Density &
\begin{minipage}[t]{\linewidth}\vspace{-6pt}
Text that is verbose but conveys little actual information \\ 
Generic statements that could apply in almost any context \\ 
Excessive filler words and phrases that add no value
\end{minipage} &
``In today's fast-paced modern world of cutting-edge technology and innovation, it has become increasingly important to consider the various factors and elements that contribute to our understanding of this complex and multifaceted issue.'' \newline \textit{(Slop because it uses many words to say almost nothing)} \\
\midrule
IU2 & Information Relevance &
\begin{minipage}[t]{\linewidth}\vspace{-6pt}
Content that fails to address the nuances of the query or task \\ 
Content that contributes nothing meaningful to context/query/task \\ 
Text that appears disconnected from its intended purpose \\ 
For text with additional context, consider relevance to such texts \\ 
For text with no additional context, consider internal relevance
\end{minipage} &
In response to ``How can I improve my marathon time?'': \newline ``Running is an excellent form of exercise with many health benefits including improved cardiovascular function, enhanced mood, and weight management.'' \newline \textit{(Slop because it doesn't address the specific question about improving marathon times)} \\
\midrule

\multicolumn{4}{l}{\textbf{\textcolor{stylequality}{Style Quality}}} \\
\midrule
SQ1 & Repetition &
\begin{minipage}[t]{\linewidth}\vspace{-6pt}
Excessive use of the same words or phrases \\ 
Redundant statements that add no new information \\ 
Overuse of transitional phrases common in formulaic writing \\ 
Low diversity in vocabulary and expression
\end{minipage} &
``The project was a success. The team accomplished their goals successfully. The successful outcome was due to the team's hard work.'' \newline \textit{(Slop due to repetition of ``success/successful'' without adding new information)} \\
\midrule
SQ2 & Templatedness &
\begin{minipage}[t]{\linewidth}\vspace{-6pt}
Over-reliance on formulaic structures and patterns \\ 
Predictable formatting patterns (e.g., excessive use of bullet points) \\ 
Standard transitional phrases used repeatedly \\ 
Frequent appearance of text following a common pattern
\end{minipage} &
``Dr. Smith, a researcher at Oxford University, found that... Professor Johnson, a scientist at Cambridge University, discovered that... Dr. Williams, an expert at Yale University, confirmed that...'' \newline \textit{(Slop because it follows the same formula repeatedly)} \\
\midrule
SQ3 & Coherence &
\begin{minipage}[t]{\linewidth}\vspace{-6pt}
Poor sentence structure or organization \\ 
Inconsistencies in argument or narrative \\ 
Text that requires significant effort to follow \\ 
How paragraphs work together to advance the argument or story
\end{minipage} &
``Climate change is affecting global temperatures. Polar bears are mammals. Ice cream melts in warm weather. Arctic ice is melting. Some people enjoy winter sports.'' \newline \textit{(Slop because the sentences, while related to temperature, don't flow logically)} \\
\midrule
SQ4 & Language Naturalness &
\begin{minipage}[t]{\linewidth}\vspace{-6pt}
Language that sounds artificial or manufactured \\ 
Strange turns of phrases or unnatural language \\ 
Technically correct grammar that still reads unnaturally \\ 
Misaligned word choice for the context \\ 
Can co-occur with verbosity if long, does not necessarily include complex words
\end{minipage} &
``The earthen area that formerly held the puddle was now dry.'' \newline \textit{(Slop because natural language would simply say ``The puddle had dried up'' or ``The ground where the puddle had been was now dry'')} \\
\midrule
SQ5 & Verbosity &
\begin{minipage}[t]{\linewidth}\vspace{-6pt}
Excessive wordiness relative to the information conveyed \\ 
Unnecessarily ``flowery'' or descriptive language \\ 
Text that prioritizes word count over meaningful content \\ 
Long-winded explanations that need significant editing
\end{minipage} &
``The consumption of the aforementioned beverage, which had been prepared with the utmost care and attention to detail by the skilled barista, provided me with a sense of satisfaction and contentment that permeated my entire being.'' \newline \textit{(Slop because it could simply say ``I enjoyed the coffee'')} \\
\midrule
SQ6 & Word Complexity &
\begin{minipage}[t]{\linewidth}\vspace{-6pt}
Inappropriate use of vocabulary relative to context \\ 
Unnecessary jargon or complicated terminology \\ 
Content filled with buzzwords that obscure meaning \\ 
Overuse of rare words
\end{minipage} &
In a general article about gardening: \newline ``The phenolic compounds in certain cultivars exhibit antimicrobial properties that mitigate pathogenic microorganism colonization.'' \newline \textit{(Slop because it uses unnecessarily complex terminology for the intended audience)} \\
\midrule
SQ7 & Tone &
\begin{minipage}[t]{\linewidth}\vspace{-6pt}
Generic voice lacking character or purpose \\ 
Missing perspective or point of view \\ 
Overly formal language in casual contexts (or vice versa) \\ 
Text that reads like an outside observer rather than engaged writer \\ 
Overconfidence in response \\ 
Can have a relationship with factuality (IQ1)
\end{minipage} &
In a blog post about personal travel experiences: \newline ``The aforementioned destination offers numerous recreational activities for tourists. Visitors may engage in swimming, hiking, or dining at local establishments.'' \newline \textit{(Slop because it uses an inappropriately formal tone for a personal blog)} \\
\end{longtable}


Careful selection of measurements to operationalize the definition of ``slop'' requires consideration of construct validity (i.e., whether we are measuring the intended phenomena), and a discussion of any errors the measurement may inadvertently introduce. 
Here, we describe the constructs comprising ``slop'' and provide an example of how to operationalize each;
we rely on prior work, where possible, for established methods of measuring each code introduced. 
Note that we aim to establish the validity of a combination of such measures to quantify ``slop'', rather than focusing on whether individual measures alone capture their intended construct.\footnote{Hence our reliance, where possible, on prior works on quantifying the individual factors considered.}  
\subsubsection*{\textcolor{infoutility}{Information Utility}}

\paragraph{Density.}
A key indicator of ``slop'' is the relatively low density of information within it. 
Such texts are often verbose without conveying much information, or contain many generic statements that are broadly applicable.
We measure information density in two ways. 
In the first, we adopt an information-theoretic approach, following \citet{Meister2021RevisitingTU}.
The uniform density hypothesis posits that speakers generally tend towards spreading information uniformly across utterances. 
We measure token entropy using GPT-2 \citep{radford2019language}, and then evaluate the mean and coefficient of variation. 
A higher mean indicates an overall lower information density in text, and a higher coefficient of variation indicates less uniformity, both of which can be indicative of ``slop''. 

In the second measurement, we measure the propositional idea density.
Ideas can be approximated by the number of verbs, adjectives, adverbs, prepositions, and conjunctions, and the density can be estimated by adjusting the counts of sets of high-likelihood part-of-speech sequences and dividing by the total number of words in a document \citep{brown2008automatic}.
Higher values of idea density indicate a higher amount of information in the text. 

\paragraph{Relevance.}
Measuring relevance (to a context), similar to factuality, is an active research area. 
Most methods assume access to a high-quality set of source documents and queries. 
Relevance, where context is provided, is measured relative to the query and task at hand. 
In the absence of additional context (e.g., task or domain), relevance can be evaluated on the internal consistency of the passage. 
``Slop'' can comprise content that fails to address the query or task, sometimes subtly.
Recently, \citet{clarke2024llm} showed that GPT-4o cannot reliably act as a replacement for human assessments of relevance for conditional generation. 
Therefore, we rely on human assessments of relevance with additional context provided where possible. 

\subsubsection*{\textcolor{infoquality}{Information Quality}}
\paragraph{Factuality.} In non-fiction texts, high-quality text is accurate. 
LLM ``slop'', however, is defined as having ``subtle inaccuracies'', introducing hallucinations (``non-existing entities''), or containing fallacious claims. 
Automatically measuring factuality is an open research problem (e.g., \citealt{ramprasad-etal-2024-evaluating, laskar2023large}), and can depend on whether reference (source) documents are available.
We rely on human annotations to detect inaccuracies in LLM-written texts in all cases.

\paragraph{Bias (Subjectivity).} Bias in text can refer to a range of topics that might influence the subjectivity of writing, and can span social \citep{Blodgett2020LanguageI} or cognitive \citep{atreides2024cognitive} facets. 
Unless explicitly prompted to produce an opinion, much of the content in ``slop'' lacks subjectivity in presenting information (factual or otherwise). 
Of the expert definitions that mention bias, there is a notable focus on the \textit{lack} of subjectivity in ``slop.'' 
There is often a missing rhetorical point of view when it is otherwise needed, or a lack of appropriate, engaged perspective. 
For instance, an LLM-generated movie review that simply states facts such as ``[...] movie received 3.5 stars and had a small budget'' does not provide any of the subjective assessments one expects in a critique. 

We use the subjectivity lexicon from \citet{wiebe2004learning}, which provides words with labels as either subjective (weak, strong) or objective. 
Following prior work, we define our bias measurement as the proportion of subjective words to total number of words in a document. 

\subsubsection*{\textcolor{stylequality}{Style Quality}}
\paragraph{Repetition.}
 When defined in the context of ``slop'', repetition entails excessive use of the same words or phrases and low diversity in vocabulary and expression. 
Prior work has looked at measuring semantic \citep{tevet-berant-2021-evaluating, namuduri2025qudsim} and lexical repetition \citep{shaib2024standardizingg}. We focus specifically on lexical repetition metrics, measuring the compression ratio over words (CR) and over parts-of-speech (CR: PoS) to capture repetitive phrases and words.

\paragraph{Templatedness.}
LLMs tend to write formulaically at the syntactic level \citep{shaib-etal-2024-detection}. 
``Slop'' may include an over-reliance on formulaic structures and patterns, such as predictable formatting (e.g., bullet points) and repeated use of certain transitional phrases. 
Following \citet{shaib-etal-2024-detection}, we measure the template rate and templates-per-token for text. 

\paragraph{Coherence and Fluency.}
Automatically measuring coherence and fluency in text is difficult \citep{li2024llms, murugadoss2024evaluatingevaluatormeasuringllms}, and may require human assessments to validate. 
Fluency is the correctness of the written language.
Coherence refers to the logical flow and connection between ideas presented in a text.
State-of-the-art LLMs that have undergone rounds of post-training and alignment rarely produce text that is completely disfluent.\footnote{At least in English; Multilingual assessments may show otherwise.} 
``Slop'', however, can exhibit low coherence (such as poor sentence organization, inconsistencies in argument or narrative, or written in a way that demands significant effort to follow), or subtle disfluencies (e.g., strange turns of phrase, technically correct but unnatural language, or word choices misaligned to the context).
We rely on expert human annotations to identify instances of disfluency or incoherence in texts.

\paragraph{Verbosity.}
LLMs tend to respond to simple queries with high verbosity, leading to training with explicit instructions to ``be concise!'' \citep{zhang2024verbosityneqveracitydemystify}.
In ``slop'', texts are often highly verbose.
We measure verbosity as the passage length (number of words), and also as the average length of sentences. 

\paragraph{Word Complexity.}
Word complexity assesses the vocabulary in a passage relative to the context: ``Slop'' can contain unnecessary jargon, buzzword-laden content, or can exhibit an overuse of ``rare'' words \citep{hovy-2016-enemy}.
Our evaluation of ``slop'' is in English texts, so we opt to use established measurements of complexity: Gunning-Fog Index, which measures the years of formal education needed to understand text on a first reading \citep{gunning1952technique}, Flesch-Kincaid Grade Level \citep{kincaid1975derivation}, measuring the (U.S.) school grade level one needs to understand the text, Flesch Reading Ease \citep{flesch1948new}, measuring textual difficulty on a 100-point scale where higher scores indicate easier-to-read text. 
We also measure sentence and word length as these directly correlate with text complexity.

\paragraph{Tone.} 
The overall tone of a text should reflect an appropriate style and voice given the context. ``Slop'' may be read as lacking character or perspective, and as containing overly formal language in casual contexts. 
This can sometimes appear as overconfidence in responses, or sycophancy \citep{fanous2025syceval, yang2024trustllmsmitigateoverconfidence}.
We rely on human annotators to identify a combination of this characterization of tone in ``slop''.

\begin{table}[t]
\centering
\resizebox{0.7\linewidth}{!}{%
\begin{tabular}{lllccc}
\toprule
\textbf{Themes} & \textbf{Final Codes} & \textbf{Pair}   & \textbf{AC$_1$} & \textbf{$\kappa$} & \textbf{Prev.\ (\%)} \\ 
\midrule
\multirow{6}{*}{\textcolor{infoutility}{\shortstack[l]{Info.\\Utility}}}
  & \multirow{3}{*}{Density}
    & A1--A2 & 0.94& 0.37  & 7.3\\
  &                      & A1--A3 & 0.23& 0.04& 49.1\\
  &                      & A2--A3 & 0.30& 0.08& 50.9\\
\cmidrule(lr){2-6}
  & \multirow{3}{*}{Relevance}
    & A1--A2 & 0.51& 0.21& 40.0\\
  &                      & A1--A3 & 0.17& 0.20& 67.3\\
  &                      & A2--A3 & 0.05& 0.09& 61.8\\
\midrule
\multirow{6}{*}{\textcolor{infoquality}{\shortstack[l]{Info.\\Quality}}}
  & \multirow{3}{*}{Factuality}
    & A1--A2 & 0.83& 0.51& 21.8\\
  &                       & A1--A3 & 0.17& 0.01& 25.5\\
  &                       & A2--A3 & 0.30& 0.02& 23.6\\
\cmidrule(lr){2-6}
  & \multirow{3}{*}{Bias}
    & A1--A2 & 0.85& 0.32& 16.4\\
  &                       & A1--A3 & 0.59& 0.02& 32.7\\
  &                       & A2--A3 & 0.75& 0.10& 21.8\\
\midrule
\multirow{9}{*}{\textcolor{stylequality}{\shortstack[l]{Style\\Quality}}}
  & \multirow{3}{*}{Structure}
    & A1--A2 & 0.47& 0.14& 41.8\\
  &                       & A1--A3 & -0.15& 0.01& 72.7\\
  &                       & A2--A3 & 0.13& 0.23& 69.1\\
\cmidrule(lr){2-6}
  & \multirow{3}{*}{Coherence}
    & A1--A2 & 0.82& 0.25& 18.2 \\
  &                       & A1--A3 & 0.36& 0.08& 45.5\\
  &                       & A2--A3 & 0.28& 0.01& 49.1\\
\cmidrule(lr){2-6}
  & \multirow{3}{*}{Tone}
    & A1--A2 & 0.47& 0.02& 40.0\\
  &                       & A1--A3 & 0.18& 0.06& 56.4\\
  &                       & A2--A3 & 0.05& -0.05& 61.8\\
\bottomrule
\end{tabular}%
}
\caption{ Gwet’s AC$_1$, Cohen’s $\kappa$, and prevalence for each annotator pair and final code.}
\label{tab:agreement-pairwise}
\end{table}

\section{Additional Data Details}
\label{app:addl-data-details}
For News, we use the articles first introduced in \citealt{Russell2025PeopleWF}: these are news articles generated by GPT-4o, Claude-3.5-
Sonnet, and O1-pro. The articles are generated by providing the title of a real news article pulled 8 American publications: Associated Press, Discover Magazine, National Geographic, New York Times, Reader’s Digest, Scientific American, Smithsonian Magazine, and Wall Street Journal. \footnote{\url{https://github.com/jenna-russell/human_detectors}}  
For MS MARCO, we randomly sample a subset of 100 passages.\footnote{\url{https://huggingface.co/datasets/microsoft/ms_marco/}} 
We filter the passages for answers longer than 30 words to ensure long enough responses for annotation. We prompt 4 models (OLMo-2-7B-Instruct, Mistral-7B, Gemma-27B, GPT4o-mini) to generate a response using the following prompt: 
\begin{quote}
    \texttt{You are given a search query and a set of potentially relevant articles. Your task is to answer the query.
    Sources: [SOURCES]
    Query: [QUERY]}
\end{quote} 

Where we replace \texttt{SOURCES} and \texttt{QUERY} with the relevant context and query from the dataset. 
For all models (where possible), we greedily generate the responses (e.g., setting the parameter for sampling to \texttt{False}). 
For open-source models, we use the HuggingFace platform to generate the text.\footnote{\url{https://huggingface.co/}}

\section{Detailed Span Statistics}
\label{app:span-stats}
Tables~\ref{app:news},~\ref{app:marco},~\ref{app:combined} provide detailed span-level statistics from the annotators over the News, MS MARCO, and combined dartasets, respectively.

\begin{table}[h]
\centering
\begin{tabular}{lcc}
\toprule
Annotator & Avg.\ Spans/Item & Avg.\ Span Length (tokens) \\
\midrule
A1 & 5.85  & 21.42 \\
A2 & 3.72  & 13.94 \\
A3 & 12.61 & 17.36 \\
\midrule
Average & 7.39 & 17.57 \\
\bottomrule
\end{tabular}
\caption{Average spans-per-item and length by annotator for the News dataset ($n=1520$ spans).}
\label{app:news}
\end{table}

\begin{table}[h]
\centering
\begin{tabular}{lcc}
\toprule
Annotator & Avg.\ Spans/Item & Avg.\ Span Length (tokens) \\
\midrule
A1 & 1.77 & 9.22  \\
A2 & 1.07 & 5.57  \\
A3 & 2.97 & 11.64\\
\midrule
Average & 1.94 & 8.81 \\
\bottomrule
\end{tabular}
\caption{Average spans-per-item and length by annotator for the MS MARCO dataset ($n=192$ spans).}
\label{app:marco}
\end{table}

\begin{table}[h]
\centering
\begin{tabular}{lcc}
\toprule
Annotator & Avg.\ Spans/Item & Avg.\ Span Length (tokens) \\
\midrule
A1 & 4.82  & 20.58 \\
A2 & 3.06  & 13.37 \\
A3 & 10.22 & 17.16 \\
\midrule
Average & 6.03 & 17.04 \\
\bottomrule
\end{tabular}
\caption{Average spans-per-item and length by annotator for the combined News and MS MARCO datasets ($n=1712$ spans).}
\label{app:combined}
\end{table}

\section{Human-LLM Span Overlap (Qualitative Assessment)}
\label{app:qual-assessment}
\begin{figure}
    \centering
    \includegraphics[width=0.8\linewidth]{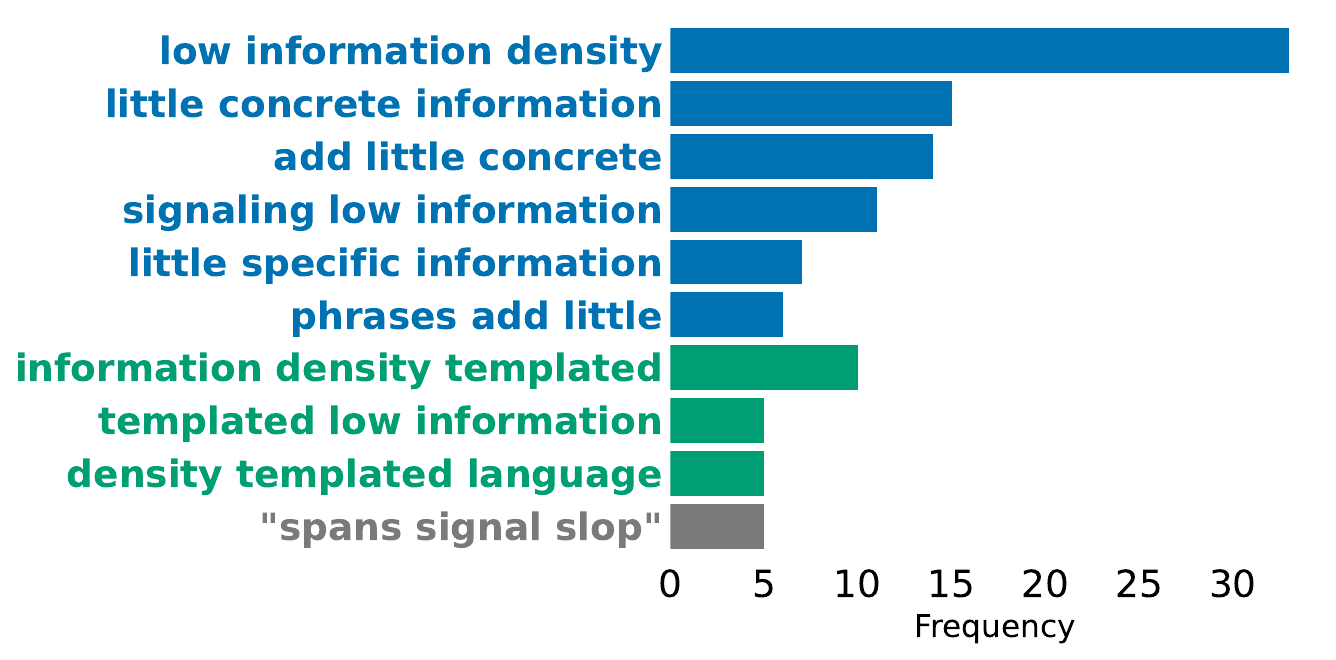}
    \caption{Tri-grams extracted from o3-mini rationales over highlighted ``slop'' spans.}
    \label{app:qual-rationale-spans}
\end{figure}

Figure~\ref{app:qual-rationale-spans} shows the top ranked trigrams, and their categorization along the ``slop'' themes by colour. o3-mini tends to overly reason about information density relative to human-assigned labels.

\section{Results by Domain and Topic}
\label{app:qualitative_domain_assessment}
\begin{figure*}[h]
    \centering
    \includegraphics[width=1.0\linewidth]{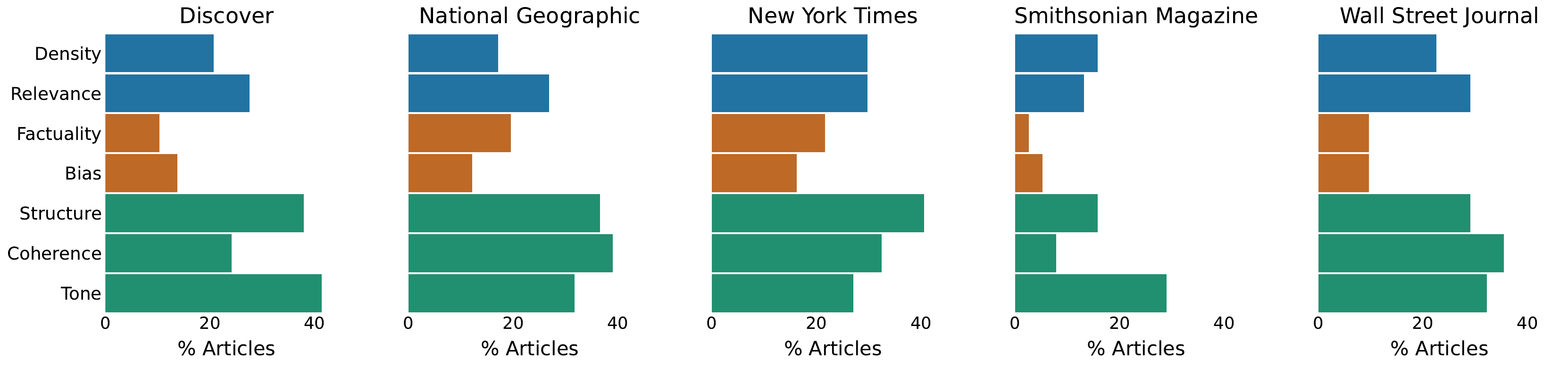}
    \caption{``Slop'' axis label prevalence stratified by news source.}
    \label{fig:source-prevalence}
\end{figure*}

In the News domain, we further assess the distribution of ``slop'' labels stratified by the source of the article (e.g., Discover, Wall Street Journal). Figure~\ref{fig:source-prevalence} shows the distribution of labels from annotators across each News source. We find that the categories are roughly represented similarly across sources (e.g., Style Quality codes annotated at a much higher rate relative to Information Quality within each source). We include the source counts in Table~\ref{tab:counts-sources}.

\begin{table}[h]
    \centering
    \begin{tabular}{lr}
        \toprule
        \textbf{Publication} & \textbf{Number} \\
        \midrule
        National Geographic & 41 \\
        Smithsonian Magazine & 38 \\
        New York Times & 37 \\
        Wall Street Journal & 31 \\
        Discover & 29 \\
        Readers Digest & 24 \\
        Associated Press & 22 \\
        Scientific American & 19 \\
        Reader's Digest & 7 \\
        \bottomrule
    \end{tabular}
    \caption{Number of articles for each News source.}
    \label{tab:counts-sources}
\end{table}\

\section{Results by Individual and by Domain}
\label{app:breakdown}
We provide the pairwise agreement among annotators for all the ``slop'' codes in Table~\ref{tab:agreement-pairwise}, including the percentage of overall prevalence of the label. A1/A2 had consistently strong agreement, whereas A2/A3 diverged. In adjudication meetings, annotators discussed these differences which can be attributed to editing style. 

\begin{figure*}[t]  
  \centering
  \begin{minipage}[t]{0.375\textwidth}
    \centering
    \includegraphics[width=\linewidth]{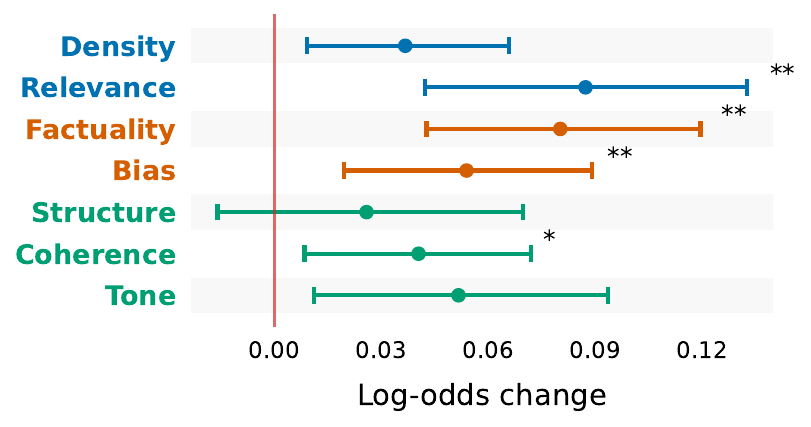}
    \subcaption{A1}
    \label{fig:a1}
  \end{minipage}
  \hspace*{\fill}
  \begin{minipage}[t]{0.30\textwidth}
    \centering
    \includegraphics[width=\linewidth]{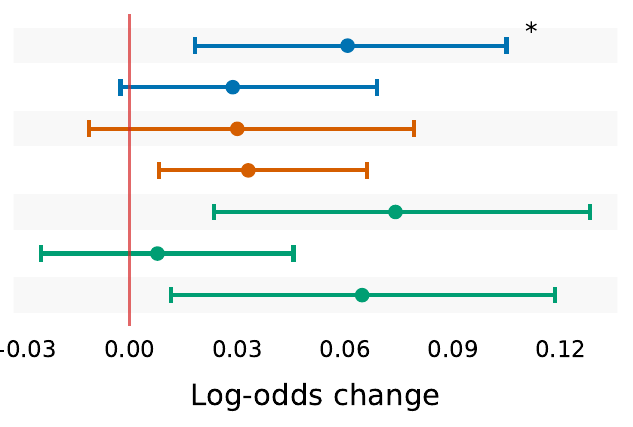}
    \subcaption{A2}
    \label{fig:a2}
  \end{minipage}
  \hspace*{\fill}
  \begin{minipage}[t]{0.30\textwidth}
    \centering
    \includegraphics[width=\linewidth]{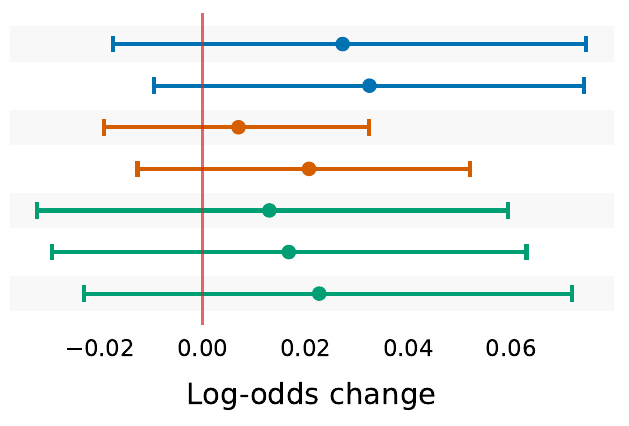}
    \subcaption{A3}
    \label{fig:a3}
  \end{minipage}

  \caption{``Slop'' codes most predictive of the overall positive label for (a) Annotator 1, (b) Annotator 2, and (c) Annotator 3 in the News domain. * $p < 0.05$, and ** $p < 0.01$. }
  \label{fig:annotator_news_split}
\end{figure*}

For all data, we find that the three annotators varied in which ``slop'' codes most strongly predicted their overall judgments (Fig.~\ref{fig:annotator_news_split}). For A1, information-related issues were more salient: Density, Relevance, Factuality, and Bias all showed strong positive associations. This suggests that A1 relied heavily on signs of low information quality or utility when identifying slop. A2, by contrast, was more selective, with verbosity (Density) emerging as the only significant predictor and Structure and Coherence showing positive though non-significant effects, indicating greater emphasis on how text was organized rather than on factual accuracy or bias. For A3, none of the codes reached significance, and while Density, Relevance, and Bias trended positive, wide confidence intervals suggest less consistency in how the taxonomy was applied. Taken together, these results highlight that annotators converge on verbosity as a core indicator of slop but diverge in how strongly they weight other dimensions such as Factuality, Bias, and Coherence.

Table~\ref{tab:agreement-news-only} shows slightly different values for the agreement metrics over the codes when looking at the News only domain. Similarly, the binary label agreement appears to be higher, with a Cohen's $\kappa$ of -0.15 (A1--A2), 0.29 (A1--A3), and 0.06 (A2--A3), indicating poor to fair agreement. 
Annotators assigned a positive label of ``slop'' to an average of 34\% of the articles. 
By contrast, Gwet's AC$_1$ yields pairwise scores of 0.12 (A1--A2), 0.42 (A1--A3), and 0.28 (A2--A3), indicating fair to moderate agreement when correcting for prevalence.  

\begin{table}[]
\centering
\resizebox{0.8\linewidth}{!}{%
\begin{tabular}{lllccc}
\toprule
\textbf{Themes} &\textbf{Final Codes}    &\textbf{$\alpha_{MASI}$}& \textbf{$\kappa$} & \textbf{AC$_1$} & \textbf{Prev. (\%)}\\ 
\midrule
\multirow{2}{*}{\textcolor{infoutility}{\shortstack[l]{Info. Utility}}} 
    &Density         &\multirow{2}{*}{\shortstack[l]{0.34}} & 0.16&  0.45&59.1\\
    &Relevance             && 0.14&  0.22&68.2\\ 
\midrule
\multirow{2}{*}{\textcolor{infoquality}{\shortstack[l]{Info. Quality}}} 
    &Factuality            &\multirow{2}{*}{\shortstack[l]{0.45}}& 0.23&  0.76&29.5\\
    &Bias                  && 0.11&  0.67&38.6\\ 
\midrule
\multirow{3}{*}{\textcolor{stylequality}{\shortstack[l]{Style Quality}}} 
    &\multirow{1}{*}{\shortstack[l]{Structure}} 
     &\multirow{3}{*}{\shortstack[l]{0.34}}& 0.11&  0.51&52.3\\
    &Coherence && 0.13&  0.39&59.1\\  
    &\multirow{1}{*}{\shortstack[l]{Tone}}
     && -0.11&  0.20&50\\
\bottomrule
\end{tabular}%
}
\caption{Agreement ($\alpha_{MASI}, \kappa, AC_{1}$) and label prevalence (Prev. \%) for ``slop'' codes in the News domain.}
\label{tab:agreement-news-only}
\end{table}

\section{Automated Metrics}
\label{app:autometrics}

\begin{table}[]
\centering
\resizebox{0.96\linewidth}{!}{
\begin{tabular}{@{}lllcl@{}}
\toprule
\textbf{Themes}                      &  \textbf{Final Codes}&\textbf{Codes}       &  \textbf{\shortstack[l]{Sig. Feature?}} &\textbf{\shortstack[l]{Auto. Metric}}\\ \midrule
\multirow{2}{*}{\textcolor{infoutility}{\shortstack[l]{Info. Utility}}} &  Density&IU1: Density         &    \greencheck &Surprisal \citep{Meister2021RevisitingTU}\\
                                     &  Relevance&IU2: Relevance       &   \greencheck &--- \\ \midrule
\multirow{2}{*}{\textcolor{infoquality}{\shortstack[l]{Info. Quality}}} &  Factuality&IQ1: Factuality      &   \greyX&--- \\
                                     &  Bias&IQ2: Bias            &    \greencheck &Subjectivity-Lexicon \citep{wiebe2004learning}\\ \midrule
\multirow{6}{*}{\textcolor{stylequality}{\shortstack[l]{Style Quality}}}       &  \multirow{2}{*}{\shortstack[l]{Structure}}&SQ1: Repetition      &    \greyX&Compression Ratios \citep{shaib2024standardizingg} \\ 
                                     &  &SQ2: Templatedness   &    \greyX&Templates-per-Token \citep{shaib-etal-2024-detection}\\\cmidrule(lr){2-5}
                                     &  Coherence&SQ3: Coherence       &    \greencheck &--- \\\cmidrule(lr){2-5}
                                     &  \multirow{4}{*}{\shortstack[l]{(Aspects of) \\Tone}}&SQ4: Fluency         &    \greyX& UniEval (partial; \citealt{li2025unievalunifiedholisticevaluation}) \\
                                     &  &SQ5: Verbosity       &    \greyX&Num. Words, Sentences, Paragraphs \\
                                     &  &SQ6: Word Complexity &    \greyX&GFI \citep{gunning1952technique}\\
                                     &  &SQ7: Tone            &   \greencheck &--- \\ \bottomrule
\end{tabular}
}
\caption{Existing automatic metrics for ``slop'' codes. We mark codes that are significant predictors of slop label with a green checkmark: 3 of 5 of significant features lack reliable automatic measures.}
\label{tab:taxonomy-auto}
\end{table}

\begin{figure*}
    \centering
    \includegraphics[width=0.7\linewidth]{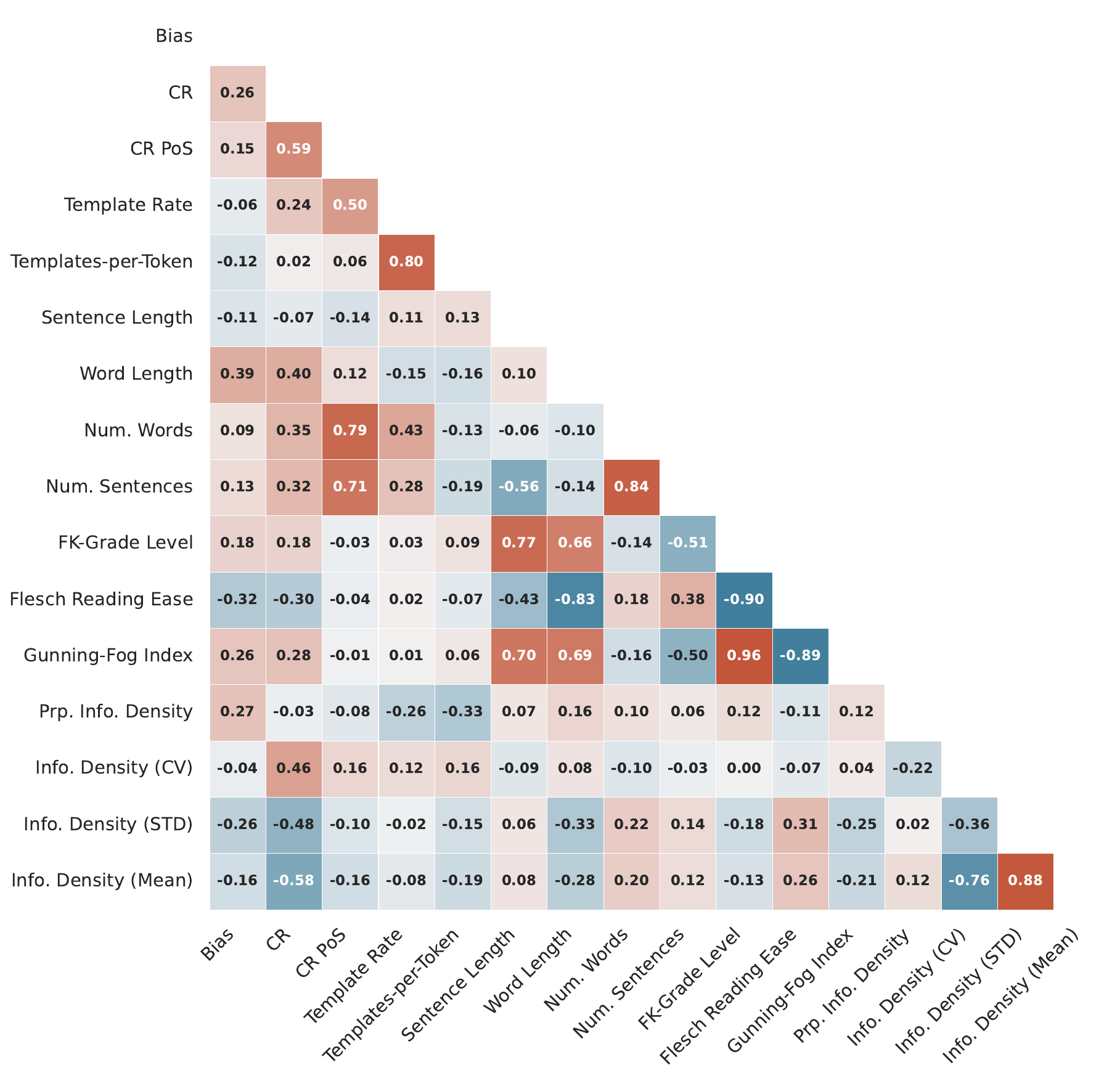}
    \caption{Correlation news}
    \label{fig:correlation}
\end{figure*}

We report the correlation between automatic text metrics in Figure~\ref{fig:correlation}. 
Many metrics have moderate to high correlations indicating shared information. 

\begin{figure*}[t]
    \centering
    \begin{subfigure}[t]{0.48\textwidth}
        \centering
        \includegraphics[width=\linewidth]{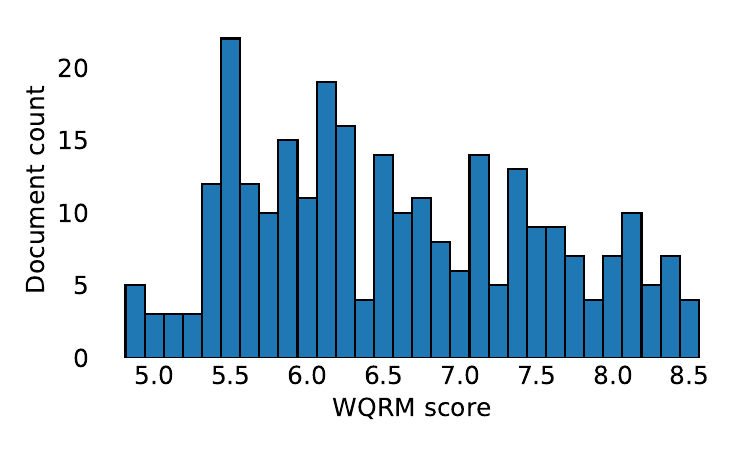}
        \label{fig:news_wqrm}
        \vspace{-1em}
        \caption{News}
    \end{subfigure}
    \hfill
    \begin{subfigure}[t]{0.48\textwidth}
        \centering
        \includegraphics[width=\linewidth]{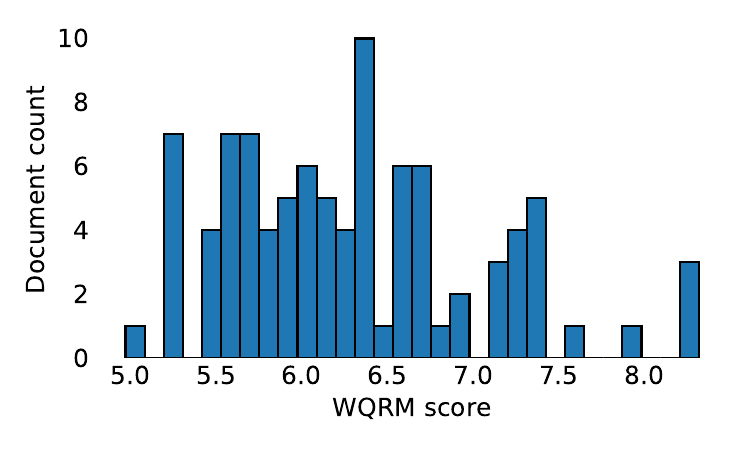}
        \label{fig:marco_wqrm}
        \vspace{-1em}
        \caption{MS MARCO}
    \end{subfigure}
    \caption{Distributions of WQRM scores across the (a) News and (b) MS MARCO datasets. }
    \vspace{-1em}
    \label{fig:wqrm}
\end{figure*}

We also report the distribution of WQRM scores in Figure~\ref{fig:wqrm} split by the (a) News and (b) MS MARCO datasets. The distribution of scores in the News domain is relatively broad. By contrast, the MS MARCO dataset shows a narrower spread, with most scores clustering between 5.5 and 7.0 and fewer documents at the extremes.

\begin{figure*}[t]
    \centering
    \begin{subfigure}[t]{0.48\textwidth}
        \centering
        \includegraphics[width=\linewidth]{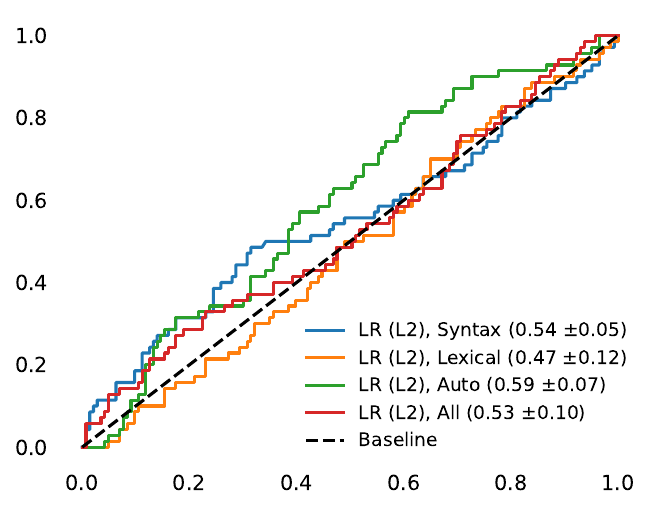}
        \label{fig:news_auroc}
        \vspace{-1em}
        \caption{News}
    \end{subfigure}
    \hfill
    \begin{subfigure}[t]{0.43\textwidth}
        \centering
        \includegraphics[width=\linewidth]{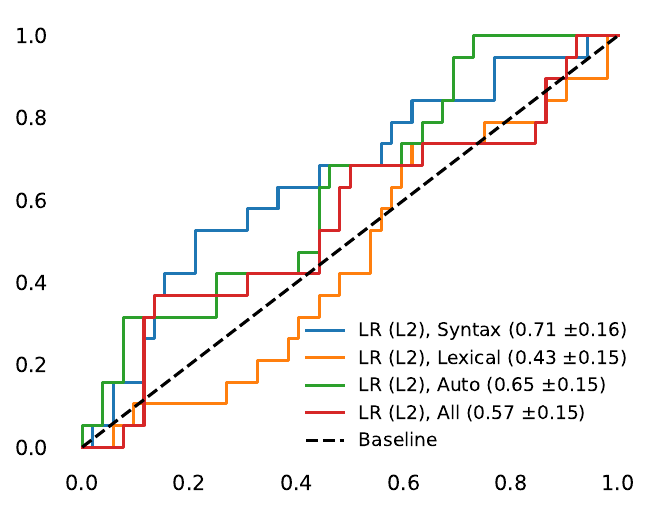}
        \label{fig:marco_auroc}
        \vspace{-1em}
        \caption{MS MARCO}
    \end{subfigure}
    \caption{AUROC using logistic regression over all available automatic text metrics (Table~\ref{tab:taxonomy-auto}) for the (a) News and (b) MS Marco datasets. In the news domain, the automatic metrics in Table~\ref{tab:taxonomy-auto} perform best, whereas the syntax feature set for MS MARCO does well, but within a wider confidence interval.}
    \vspace{-1em}
    \label{fig:disagg_auroc_LR}
\end{figure*}

\begin{figure*}[t]
    \centering
    \begin{subfigure}[t]{0.48\textwidth}
        \centering
        \includegraphics[width=\linewidth]{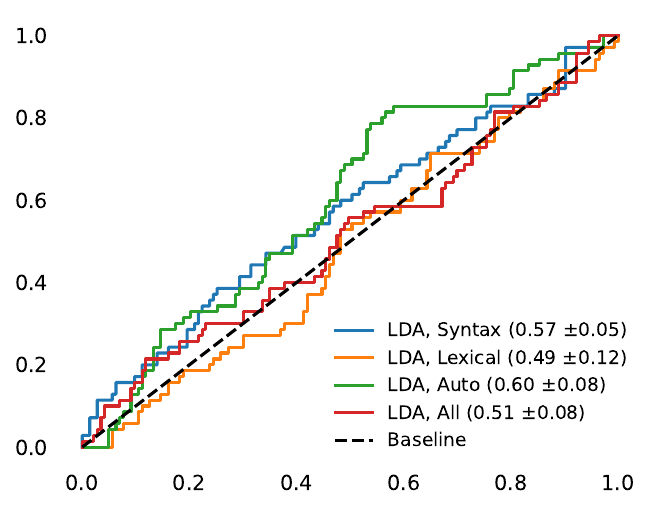}
        \label{fig:news_auroc_LDA}
        \vspace{-1em}
        \caption{News}
    \end{subfigure}
    \hfill
    \begin{subfigure}[t]{0.43\textwidth}
        \centering
        \includegraphics[width=\linewidth]{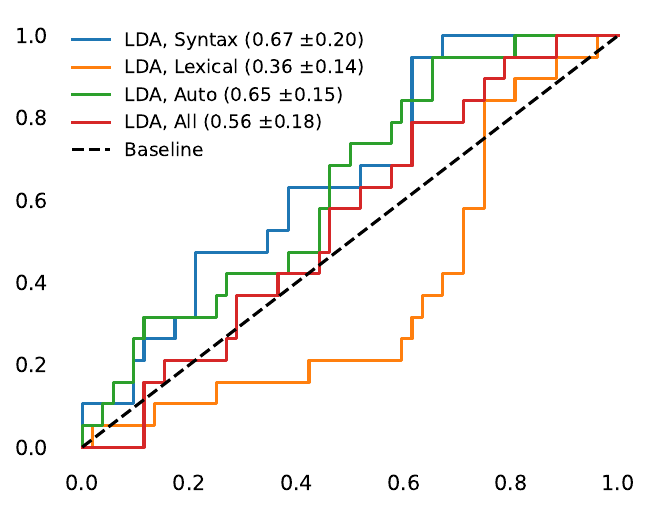}
        \label{fig:marco_auroc_LDA}
        \vspace{-1em}
        \caption{MS MARCO}
    \end{subfigure}
    \caption{AUROC using the PCA+LDA pipeline over all available automatic text metrics (Table~\ref{tab:taxonomy-auto}) for the (a) News and (b) MS Marco datasets. In the news domain, the automatic metrics in Table~\ref{tab:taxonomy-auto} perform best, whereas the syntax feature set for MS MARCO does well, but within a wider confidence interval.}
    \vspace{-1em}
    \label{fig:disagg_auroc_LDA}
\end{figure*}

We compute and show the AUROC for the automatic metrics using scikit-learn\footnote{\url{https://scikit-learn.org/stable/}} in Figures~\ref{fig:disagg_auroc_LDA},~\ref{fig:disagg_auroc_LR}. We train logistic regression models with $\ell$2 regularization using the liblinear solver. Features are standardized with a StandardScaler, and highly correlated features are removed with a threshold of 0.95. Models are tuned over a grid of 
$C\in \{0.01, 0.1, 1, 10\}$ and we balance class weights. 
For PCA+LDA, we use implementations from the scikit-learn library. We select the maximum components as the number of samples - 1, and use the SVD solver for LDA.

\begin{table}[t]
    \centering
\resizebox{0.9\linewidth}{!}{%
\begin{tabular}{lll}
\toprule
\textbf{Feature Category}& \textbf{Feature} &Description\\ 
\midrule
\multirow{5}{*}{Lexical Features}& Hapax legomena ratio&ratio of unique words to total words\\
  & Dis legomena ratio&ratio of words appearing exactly twice\\
  &  Word-length distribution&fraction of words of length 1-20\\
&  Punctuation rates&per character rates of punctuation list\\
  &  Letter frequency deviations&L2 norm of all 26 letter freq. deviations\\
\midrule
\multirow{10}{*}{Syntactic Features}&  POS tag distributions&fraction of tokens with tag from the universal SpaCy tag set\\
  &  POS tag distributions (fine-grained)&fraction of tokens with tag from the Penn Treebank\\
  &  Dependency relation distributions&fraction of tokens with dep tag\\
&  Parse tree depth&syntax tree depth per sentence (mean, std, max, median)\\
  &  Voice&ratio of passive to passive and active subjects\\
  &  Noun-phrase complexity&NP length (mean, std, max, median), and NP per sentence\\
 & Verb phrase counts&number of verb phrases\\
 & Sentence POS start&initial POS tag in sentence\\
 & Parse tree branching factor&mean and std root branching\\
 & Conjunction count&number of conjunctions\\
\bottomrule
\end{tabular}%
}
    \caption{Full list of features extracted and descriptions.}
    \label{tab:full-feature-list}
\end{table}

\section{Character-level Precision and Recall}
\label{app:pseudocode-precision}
We provide the pseudo-code for calculating character-level precision and recall for span overlap in Algorithm~\ref{algo_1}. We note that this can be modified to calculate word-level overlap, and empirically find our conclusions hold when using both character- and word-level evaluations.

\begin{algorithm}[H]
\caption{Span-level Precision}
\resizebox{0.8\linewidth}{!}{%
\begin{minipage}{\linewidth}
\begin{algorithmic}[1]

\For{\text{each document} $i$}
\State $\text{text} \gets i.\text{text}$
\\
\For{\text{each annotator} $A$}

\State $\text{spans}_A \gets \text{all highlighted spans for } i$
\State $\text{chars}_A \gets \text{char start and end for each span}$
\EndFor
\\
\For{\text{each annotator pair} $(A_1, A_2)$} 
\If{$\text{chars}_{A1}$ and $\text{chars}_{A2}$ are empty} 
\State $\text{precision}_{A1, A2} \gets 1.0$ 
\Else 
\State $\text{precision}_{A1, A2} \gets  \frac{|\text{chars}_{A1}|\  \cup \ |\text{chars}_{A2}|}{|\text{chars}_{A1}|} $ \Comment{if $\text{chars}_{A1}$ is 0, set to 0.0}
\EndIf
\\
\State \text{Record $\text{precision}_{A1, A2}$} for $i$
\EndFor
\\
\EndFor
\\
\State \Return precision
\end{algorithmic}
\end{minipage}
}
\label{algo_1}
\end{algorithm}

\section{Human-LLM Span Overlap (Qualitative Assessment)}
\label{app:qualitative_domain_assessment_human}
We investigate the \textit{word}-level overlap between the spans highlighted by o3-mini and annotators (Table~\ref{tab:f1s}) and find that there is both low precision and recall of the problematic spans identified by the reasoning model and humans. This alone indicates a misalignment between the identified slop spans, however we also compare the label distributions between the two as suggested and find that, qualitatively, the model tends to favour explanations aligned with the Density category, but misses other axes frequently identified by humans (e.g., Relevance, Coherence, Tone).

\begin{table}[h]
    \centering
    \begin{tabular}{ll}
        \toprule
        \textbf{Metric} & \textbf{Value} \\ 
        \midrule
        precision & 0.14 \\ 
        recall & 0.13 \\ 
        f1 & 0.14 \\ 
        \bottomrule
    \end{tabular}
    \caption{Word-overlap metrics between o3-mini and human span  annotations annotations.}
    \label{tab:f1s}
\end{table}

We also notice that while increasing the shots guides the reasoning to qualitatively align better with the human-written responses, the actual spans selected remain fairly similar (word-level overlap) even when providing examples in context. We evaluate the overlap of spans selected by the model with 0 and 5 examples given in context, with precision$=0.65$, recall$=0.61$, and f1=$0.63$ indicating very similar highlights regardless of in-context examples.

\section{LLM-Generated Span Rationales}
\label{app:qualitative_domain_assessment_llm}
\begin{figure}[t]
\centering
\resizebox{0.85\textwidth}{!}{
\begin{subfigure}{0.48\textwidth}
    \centering
    \includegraphics[width=\textwidth]{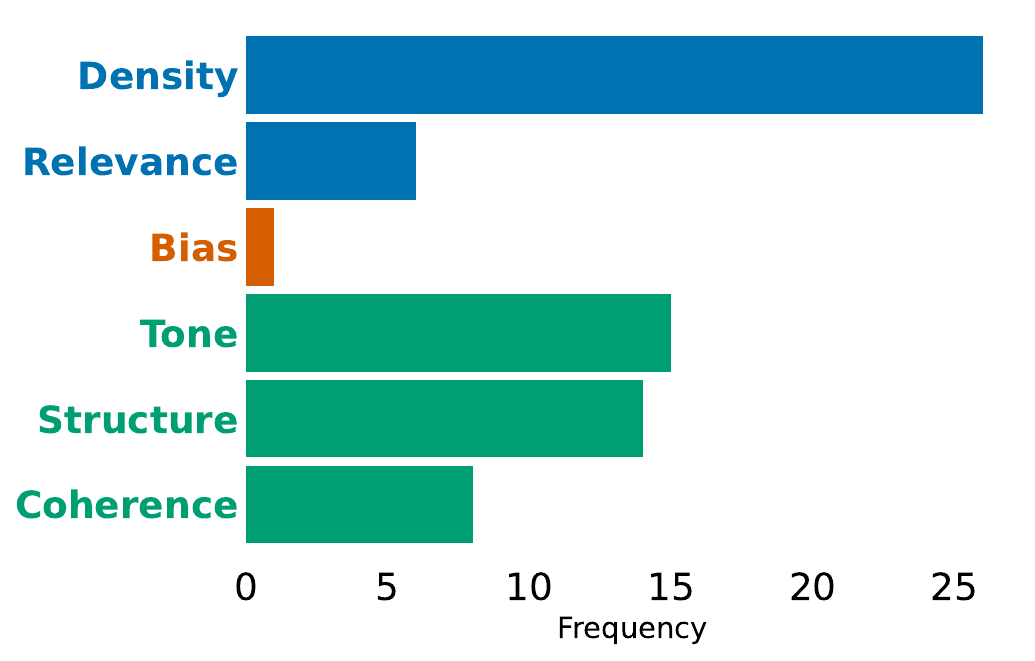}
    \caption{}
    \label{fig:qual_human}
\end{subfigure}
\begin{subfigure}{0.48\textwidth}
    \centering
    \includegraphics[width=\textwidth]{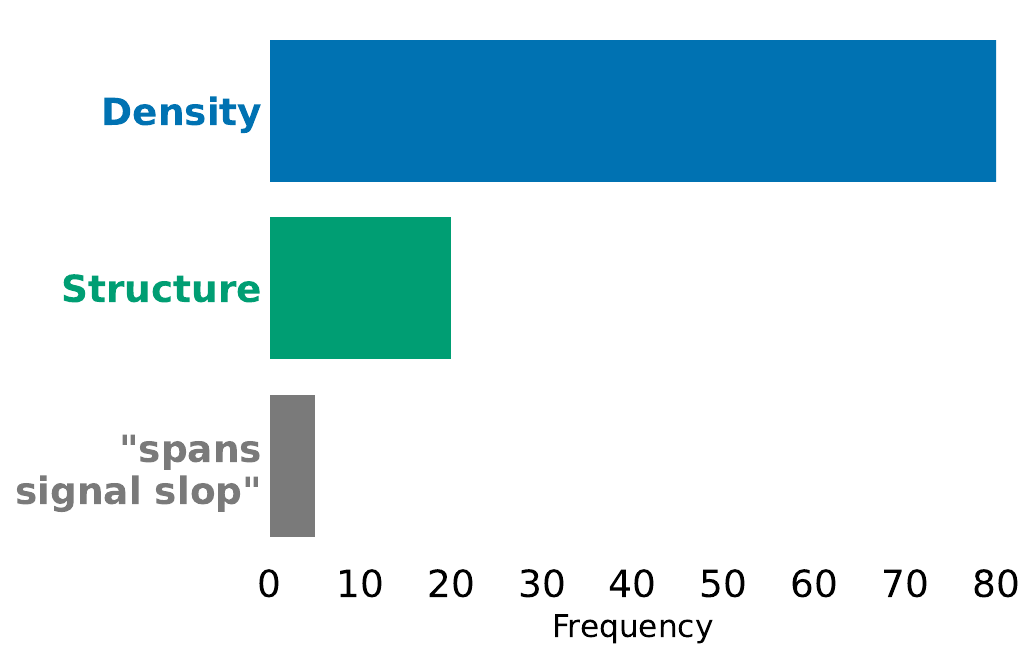}
    \caption{}
    \label{fig:qual_llm}
\end{subfigure}
}
\caption{Frequency of (a) human-assigned slop code prevalence and, (b) ``slop'' category collapsed trigrams in o3-mini span rationales in the News domain. o3-mini over-represents issues related to Density in the rationales, while under-representing issues with Coherence, Tone, Bias, Relevance, and Factuality.}
\label{fig:qual-rationale-spans}
\end{figure}

In addition to extracting spans, the LLM-as-a-judge is instructed to provide a rationale for each selection. 
Reasoning chains do not reliably return the exact codes from the guide, so to assess the overall alignment of the reasoning and the human-assigned codes, we count and rank tri-grams from the judge-generated rationale (Figure~\ref{fig:qual-rationale-spans}, Appendix Figure~\ref{app:qual-rationale-spans}). Figures~\ref{fig:qual_human} and \ref{fig:qual_llm} highlight a mismatch between the human-assigned codes and those flagged most frequently by o3-mini. Specifically, o3-mini overwhelmingly focuses on Density-related issues, and does not show the full range of codes used by annotators. In Appendix~\ref{app:qual-assessment}, we show that this does not significantly change when in-context examples are provided to the model.





\section{LLM Evaluations}
\label{app:llm-evals}

\subsection{Training}
We trained Qwen-7B-reasoning for 5 epochs. We used a learning rate of $2\times 10^{-4}$ with bf16 precision. To address class imbalance, we applied a positive oversampling rate of 0.5. 
We used the following prompt to guide answers during training: 

\begin{quote}  
\texttt{You are a careful copy editor. Given a paragraph, extract the minimal set of short verbatim spans (quoted) that are indicative of slop according to the guide, then provide a brief reasoning. The guide is provided below. Slop refers to AI-generated text that is low-quality. It can appear superficially correct but is some combination of generic, overly verbose, inaccurate, irrelevant to its intended purpose, and contributing little meaningful value to the reader (despite sounding fluent). Slop typically displays patterns of repetition, formulaic structure, vague language, and an absence of authentic perspective. \\\\
\textbf{[[Truncated]]} \\\\
Return a JSON ONLY, no prose, with keys exactly as follows: \\
\{spans: ..., reasoning: ...\}}
\end{quote}

where the \texttt{\textbf{[[Truncated]]}} section has a copy pasted version of the codes and their examples/definitions (Fig.~\ref{app:code-defn}).

To extract the silver-label rationales, we use the following prompt on o4-mini reasoning models to gather reasoning chains: 

\begin{quote}
\texttt{SYSTEM\_PROMPT = """\\
You are an experienced copy-editor.\\
For each numbered span you receive, write **one sentence ($\leq$ 25 words)**\\
explaining why the span is low-quality "slop," using its FINAL CODE as the label.\\
Return the rationales in exactly the same numbered order—nothing else.\\
""".strip()}
\\
\texttt{SLOP\_GUIDE = """\\
"Slop" = AI-generated text that is generic, verbose, inaccurate, irrelevant, or\\
adds little real value. It often shows repetition, formulaic structure, vague\\
language, and no authentic perspective.\\
FINAL CODES (7-way collapse)\\
• Density – Many words, little information; filler or fluff.\\
• Relevance – Off-topic or tangential to the passage/question.\\
• Factuality – Incorrect, fabricated, or misleading statement.\\
• Bias – One-sided, over-general, or unnuanced claim.\\
• Structure – Repetitive or templated sentence / formula pattern.\\
• Coherence – Disjointed or ill-logical flow; hard to follow.\\
• Tone – Awkward fluency, needless jargon, verbosity, or style unsuited\\
\hspace*{1em}to context/audience.\\
""".strip()}

\texttt{TASK = """
Give numbered rationale(s) ($\leq$ 25 words) per span. First, state the span label, then the rationale.\\
Output **only** the rationale list—no extra commentary as a python LIST. 
""".strip()}
\end{quote}
\begin{table}[t]
\centering
\resizebox{0.6\textwidth}{!}{
\begin{tabular}{ll}
\toprule
\textbf{LAMP Category} & \textbf{Slop Taxonomy} \\
\midrule
Cliche & Tone \\
Poor Sentence Structure & Coherence \\
Awkward Word Choice and Phrasing & Tone \\
Tense Inconsistency & Tone \\
Unnecessary/Redundant Exposition & Density, Repetition \\
Lack of Specificity and Detail & Relevance \\
\bottomrule
\end{tabular}
}
\caption{Mapping of the categories in 
\citealt{chakrabarty2025ai} to the ``slop'' taxonomy.}
\label{tab:mapping}
\end{table}

\subsection{Data Augmentation (LAMP)}
For the LAMP data \cite{chakrabarty2025ai}, we first filter for text in either the Travel Writing, Food Writing, or Creative Non-Fiction categories to match our News and QA data settings. We then map the labels  map the following categories to our ``slop'' taxonomy (Table~\ref{tab:mapping}).

\subsection{Prompting}
For prompting off-the-shelf GPT and DeepSeek models in zero- and few-shot settings, we used the following prompt(s). 

\begin{quote}
\small
\texttt{SYSTEM\_PROMPT\_SPANS = (\\
\hspace*{1em}"You are a careful copy editor. Given a paragraph, extract the minimal set of short "\\
\hspace*{1em}"verbatim spans (quoted) that are indicative of \textbackslash"slop\textbackslash" according to the guide, then provide a brief reasoning.\textbackslash n"\\
\hspace*{1em}"The guide is provided below. "Slop" refers to AI-generated text that is low-quality. It can appear superficially correct but is some combination of generic, overly verbose, inaccurate, irrelevant to its intended purpose, and contributing little meaningful value to the reader (despite sounding fluent). "Slop" typically displays patterns of repetition, formulaic structure, vague language, and an absence of authentic perspective."\\
\hspace*{1em}"Factuality: Incorrect or fabricated information, Misleading or fallacious claims. Example: "Dr. Sarah Johnson of Harvard University published groundbreaking research on this topic in 2022." (Slop if Dr. Johnson doesn't exist, isn't at Harvard, or didn't publish such research)"\\
\hspace*{1em}"Bias: Lack of appropriate perspective or over-standardization. Example: "The economic policy changes of 2023 were universally beneficial." (Slop because it presents a one-sided view of complex policy impacts)"\\
\hspace*{1em}"Information Density: Text that is verbose but conveys little actual information. Excessive filler words. Example: "In today's fast-paced modern world of cutting-edge technology and innovation, it has become increasingly important to consider the various factors and elements that contribute to our understanding of this complex and multifaceted issue." (Slop because it uses many words to say almost nothing)"\\
\hspace*{1em}"Information Relevance: Appropriateness to the specific context, query, or task. For text with no additional context (e.g., an article), consider internal relevance within the passage. Example: In response to "How can I improve my marathon time?": "Running is an excellent form of exercise with many health benefits including improved cardiovascular function, enhanced mood, and weight management." (Slop because it doesn't address the specific question about improving marathon times)"\\
\hspace*{1em}"Repetition: Excessive use of the same words or phrases. Low diversity in vocabulary and expression. Example: "The project was a success. The team accomplished their goals successfully. The successful outcome was due to the team's hard work." (Slop due to repetition of "success/successful" without adding new information)"\\
\hspace*{1em}"Templatedness: Over-reliance on formulaic structures and patterns. Predictable formatting patterns (e.g., excessive use of bullet points). Frequent appearance of text that follows a common pattern (e.g., "Mr. X, a Y-year-old Z"). Example: "Dr. Smith, a researcher at Oxford University, found that… Professor Johnson, a scientist at Cambridge University, discovered that… Dr. Williams, an expert at Yale University, confirmed that…" (Slop because it follows the same formula repeatedly)"\\
\hspace*{1em}"Coherence: Poor sentence structure or organization. Text that requires significant effort to follow. Example: "Climate change is affecting global temperatures. Polar bears are mammals. Ice cream melts in warm weather. Arctic ice is melting. Some people enjoy winter sports." (Slop because the sentences, while related to temperature, don't flow logically)"\\
\hspace*{1em}"Fluency: Strange turns of phrases or unnatural language. Example: "The earthen area that formerly held the puddle was now dry." (Slop because natural language would simply say "The puddle had dried up" or "The ground where the puddle had been was now dry")"\\
\hspace*{1em}"Word Complexity: Unnecessary jargon or complicated terminology. Overuse of rare words. Example: In a general article about gardening: "The phenolic compounds in certain cultivars exhibit antimicrobial properties that mitigate pathogenic microorganism colonization." (Slop because it uses unnecessarily complex terminology for the intended audience)"\\
\hspace*{1em}"Tone: Appropriate voice and style for the context. Example: In a blog post about personal travel experiences: "The aforementioned destination offers numerous recreational activities for tourists. Visitors may engage in swimming, hiking, or dining at local establishments." (Slop because it uses an inappropriately formal tone for a personal blog)"\\
\hspace*{1em}'Return a JSON object: \{ "spans": ["...","..."], "reasoning": "..." \}'\\
)}

\texttt{SYSTEM\_PROMPT\_LABEL = (\\
\hspace*{1em}"You are a careful copy editor. Given a piece of text, return a binary assessment of whether this is overall slop. "\\
\hspace*{1em}"The guide is provided below. "Slop" refers to AI-generated text that is low-quality. It can appear superficially correct but is some combination of generic, overly verbose, inaccurate, irrelevant to its intended purpose, and contributing little meaningful value to the reader (despite sounding fluent). "Slop" typically displays patterns of repetition, formulaic structure, vague language, and an absence of authentic perspective."\\ \\
\hspace*{1em}\textbf{[[... slop codes and examples ...]]}\\
\\
Task: Is this slop (0 = no, 1 = yes)}
\end{quote}

Where \texttt{\textbf{[[... slop codes and examples ...]]}} has the full slop guide (definitions and examples) formatted into the text.
\subsection{Single-Code LLM-as-a-Judge}
We additionally test GPT-5 as a judge for each code in the ``slop'' taxonomy. Rather than providing the full guide, we split judge evaluations by code. For example, we call the model using the guide section corresponding to a single code, resulting in multiple calls per article.

\paragraph{Annotating Behaviours}Table~\ref{app:span-length-single-code} shows the individual span counts and length compared to the human counts. We additionally count the most frequent 4-8 grams in the rationales provided by GPT-5, and find that a majority of the rationales focus on factual inaccuracies, jargon, and clarity (Table~\ref{app:single-code-qual}). 

\begin{table}[]
\centering
\begin{tabular}{llccc}
\toprule
Theme & Source & Avg Spans & Any Rate & Avg Span Length \\
\midrule
\multirow{2}{*}{Bias}       & Human & 0.14 & 0.14 & 98.67  \\
                            & GPT-5 & 1.55 & 0.91 & 87.27  \\
\cmidrule(lr){1-5}
\multirow{2}{*}{Coherence}  & Human & 2.46 & 0.32 & 132.63 \\
                            & GPT-5 & 1.55 & 0.82 & 82.50  \\
\cmidrule(lr){1-5}
\multirow{2}{*}{Density}    & Human & 0.46 & 0.32 & 134.50 \\
                            & GPT-5 & 2.00 & 1.00 & 101.09 \\
\cmidrule(lr){1-5}
\multirow{2}{*}{Factuality} & Human & 0.50 & 0.23 & 89.23  \\
                            & GPT-5 & 1.41 & 0.73 & 84.07  \\
\cmidrule(lr){1-5}
\multirow{2}{*}{Relevance}  & Human & 0.73 & 0.27 & 80.00  \\
                            & GPT-5 & 1.68 & 0.91 & 109.46 \\
\cmidrule(lr){1-5}
\multirow{2}{*}{Structure}  & Human & 1.32 & 0.55 & 293.80 \\
                            & GPT-5 & 1.34 & 0.68 & 55.34  \\
\cmidrule(lr){1-5}
\multirow{2}{*}{Tone}       & Human & 1.79 & 0.56 & 141.75 \\
                            & GPT-5 & 1.48 & 0.77 & 74.56  \\
\bottomrule
\end{tabular}
\caption{Individual span counts, appearance, and length for each individual ``slop'' code. 
}
\label{app:span-length-single-code}
\end{table}


\begin{table}[t]
\centering
\begin{tabular}{lrr}
\toprule
Text & Count & n \\
\midrule
the article stays focused                  & 56 & 4 \\
no clear factual inaccuracies              & 49 & 4 \\
and easy to follow                          & 44 & 4 \\
no unnecessary jargon buzzwords            & 34 & 4 \\
appropriate word choice no awkward grammar & 38 & 6 \\
no awkward grammar or unnatural turns      & 32 & 6 \\
relies on stock transitional phrases and   & 21 & 6 \\
winded and in need of tightening           & 19 & 6 \\
the article follows a clear chronological  & 11 & 6 \\
\bottomrule
\end{tabular}
\caption{N-grams and counts of reasoning provided the GPT-5 in the single-code span extraction setting.}
\label{app:single-code-qual}
\end{table}

\subsection{Text Quality Reward Models} 
\label{app:wqrm}
We evaluate models trained elsewhere for writing quality. We use the Writing Quality Reward Model (WQRM; \citealt{chakrabarty2025ai}) to assign quality scores to our data. 
The model is trained on paragraph-level annotations, so we split all our News data into paragraphs.

\textbf{Results.} Figure~\ref{fig:wqrm} shows the distribution of WQRM scores over the News and MS MARCO datasets, ranging from around 5.0 to 8.5, with most documents clustering between 5.5 and 7.5, indicating medium to moderately-high quality writing \citep{chakrabarty2025ai}. 
Correlations with our slop annotations demonstrates that WQRM aligns with, but does not fully capture the ``slop'' label. WQRM shows lower correlation with the binary ``slop'' label: 0.25 for News, 0.15 for MS MARCO, suggesting it captures some signal of lower writing quality. When evaluating the number of annotated spans to the WQRM, correlation is 0.48 for News and 0.19 for MS MARCO, suggesting that as the number of issues in a text increases, writing quality decreases. These results indicate that WQRM captures some axes of ``slop'', especially in settings with multiple annotated issues.

\subsection{Training Span Extraction Models}
We trained a Qwen-7B reasoning model (DeepSeek-R1-Distill-Qwen-7B; \citealt{deepseekai2025deepseekr1incentivizingreasoningcapability}) for slop span extraction. To provide rationales, we first generated silver annotations by prompting GPT-5 with the annotated span and label, asking for concise explanations of the label. We  also augment our News and MS Marco data with data from LAMP \citep{chakrabarty2025ai,10.1145/3706598.3713559}, mapping their categories (such as cliche, redundant/exposition) into our slop taxonomy to create a consistent label space. We filter the LAMP dataset to remove the creative writing subsets. We provide details of prompts used, label mappings, and training details in Appendix~\ref{app:llm-evals}.

\textbf{Results.}
Evaluation on held-out data shows that the model achieves partial-overlap (character-level) scores of 0.33 precision, 0.22 recall, and 0.26 F1. Restricting to positive-only examples results in an F1 of 0.30 (precision 0.48, recall 0.22). The model also learns to abstain from predicting spans where there was no slop (similar to annotators), with an empty prediction rate of 44\%. These results suggest that while the model can extract some slop spans with reasonable precision, it is still difficult to identify all issues.
Training a model results in better performance than prompting GPT-5 with the guide, but neither perform especially well. 
This indicates a need for more research into metrics for identifying ``slop'' spans in texts. 
\end{document}